\documentclass{article}

\usepackage{microtype}
\usepackage{graphicx}
\usepackage{subfigure}
\usepackage{booktabs} 

\usepackage{hyperref}

\usepackage[accepted]{icml2023}

\usepackage{amsmath}
\usepackage{amssymb}
\usepackage{mathtools}
\usepackage{amsthm}

\usepackage[capitalize,noabbrev]{cleveref}

\usepackage{enumitem}
\renewcommand{\algorithmiccomment}[1]{\bgroup\hfill \(\triangleright\) #1\egroup}
\usepackage{multirow}

\theoremstyle{plain}

\theoremstyle{definition}

\theoremstyle{remark}

\usepackage[textsize=tiny]{todonotes}

\icmltitlerunning{Transformer-based Stagewise Decomposition for Large-Scale Multistage Stochastic Optimization}

\begin{document}

\twocolumn[
\icmltitle{Transformer-based Stagewise Decomposition \\ for Large-Scale Multistage Stochastic Optimization}



\icmlsetsymbol{equal}{*}

\begin{icmlauthorlist}
\icmlauthor{Chanyeong Kim}{equal,yyy}
\icmlauthor{Jongwoong Park}{equal,comp}
\icmlauthor{Hyunglip Bae}{yyy}
\icmlauthor{Woo Chang Kim}{yyy}
\end{icmlauthorlist}

\icmlaffiliation{yyy}{Department of Industrial and Systems Engineering, KAIST, Daejeon, Republic of Korea}
\icmlaffiliation{comp}{NCSOFT Corporation, Seongnam, Republic of Korea}

\icmlcorrespondingauthor{Woo Chang Kim}{wkim@kaist.ac.kr}

\icmlkeywords{Machine Learning, ICML}

\vskip 0.3in
]



\printAffiliationsAndNotice{\icmlEqualContribution} 

\begin{abstract}
Solving large-scale multistage stochastic programming (MSP) problems poses a significant challenge as commonly used stagewise decomposition algorithms, including stochastic dual dynamic programming (SDDP), face growing time complexity as the subproblem size and problem count increase.
Traditional approaches approximate the value functions as piecewise linear convex functions by incrementally accumulating subgradient cutting planes from the primal and dual solutions of stagewise subproblems.
Recognizing these limitations, we introduce TranSDDP, a novel Transformer-based stagewise decomposition algorithm.
This innovative approach leverages the structural advantages of the Transformer model, implementing a sequential method for integrating subgradient cutting planes to approximate the value function.
Through our numerical experiments, we affirm TranSDDP's effectiveness in addressing MSP problems.
It efficiently generates a piecewise linear approximation for the value function, significantly reducing computation time while preserving solution quality, thus marking a promising progression in the treatment of large-scale multistage stochastic programming problems.
\end{abstract}
\section{Introduction}
Sequential decision-making problems under uncertainty can be addressed using various methodologies, including stochastic optimal control (SOC), reinforcement learning (RL), and multistage stochastic programming (MSP). 
SOC is a mathematical optimization approach that focuses on determining an optimal control policy within a continuous-time framework.
On the other hand, RL and MSP are suitable for discrete-time problems.
RL utilizes simulated or real-world experience to learn optimal decision-making through a trial-and-error process, while MSP seeks to find optimal decisions through mathematical formulation in either continuous or discrete spaces.
Therefore, the selection of a methodology depends on specific problem characteristics, such as the nature of uncertainties and the type of decision variables.

In this paper, we focus on the use of MSP to solve sequential decision-making problems under uncertainty. 
MSP applies to a wide range of problems, including production optimization, hydropower production planning, and asset liability management \citep{shiina2003multistage, carino1994russell, fleten2008short}. 
To handle uncertainties, MSP utilizes scenario trees that approximate the underlying stochastic process by a finite set of scenarios with discretized time points.
This enables the transformation of the original stochastic problem into a single large deterministic equivalent problem.
However, the computational complexity of solving these deterministic equivalent problems can become intractable as the number of scenarios increases exponentially with the number of stages and/or the number of nodes per stage.
This challenge, known as the curse of dimensionality, presents a critical limitation of MSP when confronted with large-scale problems.

To tackle the computational complexities arising from large-scale MSP problems, several decomposition-based algorithms have been proposed. 
These algorithms can be broadly categorized into two types: scenario decomposition and stagewise decomposition. 
Both approaches aim to decompose the original problem into smaller subproblems that are computationally tractable, while still ensuring adherence to the nonanticipativity constraints, which prevent decision-making from relying on future information. 

Scenario decomposition algorithms decompose large-scale MSP problems into smaller subproblems based on scenarios, allowing for the relaxation of nonanticipativity constraints within each subproblem. 
Well-known scenario decomposition algorithms include the dual decomposition algorithm \citep{caroe1999dual} and progressive hedging \citep{rockafellar1991scenarios}, which have been widely used to solve various large-scale stochastic programs \citep{triki2005optimal, berg2014optimal, fadda2019progressive}. 
However, these methods are inefficient for extremely large-scale MSP problems because they require considering every scenario at every iteration to obtain solutions that satisfy the nonanticipativity constraints after combining the subproblems.

Stagewise decomposition algorithms decompose large-scale MSP problems into a series of smaller subproblems that are solved sequentially. 
These algorithms use the value function to capture the impact of immediate decisions on future stages within each subproblem.
However, it can be challenging to find the exact explicit form of the value function, even for simple problems.
Hence, the primary objective of stagewise decomposition techniques is to attain precise and efficient approximations of the value function.
The most popular stagewise decomposition algorithm is stochastic dual dynamic programming (SDDP) \citep{pereira1991multi}, which is an extension of nested Benders decomposition. 
In SDDP, the value function is approximated using a piecewise linear convex function.
As the algorithm iterates, the piecewise linear convex function is updated by adding subgradient cutting planes which are constructed gradually based on the primal and dual solution of each subproblem.
While SDDP is guaranteed to converge to the optimal solution under mild conditions, its computational time increases at each iteration due to the monotonically growing size of the piecewise linear approximation.

In recent years, there has been significant research dedicated to improving the efficiency of the SDDP algorithm. 
These studies can be broadly categorized into three main areas: parametric value function approximation, selection of subgradient cutting planes, and generation of subgradient cutting planes.
One notable recent advancement in the field of parametric value function approximation is the value function gradient learning (VFGL) algorithm proposed by \citet{LEE2022}, which approximates the value function as a fixed parametric function form.
VFGL algorithm utilizes gradient information to optimize the parameters of a parametric function.
Due to the value function being approximated using a fixed parametric form, the size of the subproblems in each iteration remains almost constant, leading to computational stability.
However, ensuring the quality of the solution heavily relies on selecting a suitable parametric form \citep{bae2023deep}.
Selection of subgradient cutting planes is a strategy that aims to decrease the computational burden of the algorithm by constructing a piecewise linear lower bound for the value function using cutting planes that satisfy certain heuristic conditions \citep{pfeiffer2012two, de2015improving}.
This mechanism yields a substantial reduction in the subproblem sizes, leading to enhanced algorithmic efficiency.
However, the methods introduced above for solving large-scale MSP problems suffer from the drawback that even a slight perturbation in the problem necessitates solving it anew.
To handle this limitation, the process of generating subgradient cutting planes involves utilizing a neural network to acquire the capability of generating subgradient cutting planes tailored to a family of problem instances.
This enables the instantaneous generation of cutting planes for problem instances belonging to the same family, obviating the necessity of explicit construction.
Our study belongs to the category of generation of subgradient cutting planes, and the current novel and representative study of this approach is $\nu$-SDDP \citep{dai2021neural}. 
In this study, a multi-layer perceptron is trained using meta-learning to generate a fixed number of corresponding subgradient cutting planes based on the problem context vector. 
However, this approach has some limitations.
Specifically, previously generated cutting planes are not considered when generating new cutting planes due to utilizing only the problem context vector as input for the neural network.
Furthermore, $\nu$-SDDP is limited to linear programs and produces a fixed number of cutting planes.

The process of sequentially computing subgradient cutting planes in SDDP suggests the potential for applying sequence models, such as the Transformer \citep{vaswani2017attention}, to the generation of subgradient cutting planes. 
The Transformer model has achieved state-of-the-art performance across various domains, including machine translation, speech recognition, DNA sequence analysis, and video activity recognition \citep{wang2019learning, dong2018speech, ji2021dnabert, girdhar2019video}. 
Additionally, it has been successfully applied to solve optimization problems that involve sequential decision-making, such as the traveling salesman problem \citep{kool2018attention}.

Based on this insight, we propose TranSDDP which uses Transformer to generate the piecewise linear function for approximating the value function in SDDP. 
While there has been active research in RL on approximating value functions using Transformer, as demonstrated in works \citep{parisotto2020stabilizing, chen2021decision, liu2023constrained}, our study introduces the first investigation of Transformer-based value function approximation in the domain of MSP.
Our proposed model employs a sequential approach to approximate the value function by utilizing a Transformer-based architecture for generating subgradient cutting planes.
The encoder takes in vectors representing the parameters of stochastic elements, and the decoder generates a piecewise linear lower bound of the value function.
This approach yields notable improvements in the efficiency of SDDP, particularly for large-scale problems.
Moreover, as the network learns to generate cutting planes for a family of problems, it can approximate the value function for new problems without requiring problem-solving from scratch.
Furthermore, our model incorporates previously generated subgradient cutting planes into the generation of new ones by leveraging Transformer's advantage, thereby alleviating the restrictions of $\nu$-SDDP.
We demonstrate that the TranSDDP can learn to generate subgradient cutting planes without compromising the quality of the solutions.

The structure of the paper is as follows.
Section 2 introduces the problem formulations and provides an overview of the background related to SDDP, VFGL, heuristic-conditioned SDDP, and Transformer.
In section 3, we present the derivation of TranSDDP.
Section 4 discusses the numerical experiments conducted to evaluate the proposed model.
Finally, Section 5 concludes the paper, summarizing the findings and discussing potential future works.
\section{Preliminary}

In this section, we begin by providing a formal definition of the problem setting for the MSP problem.
Subsequently, we introduce an overview of SDDP, VFGL, and $\nu$-SDDP, which serve as benchmark methods in our study.
Furthermore, we present an explanation of the Transformer model that is employed in our proposed algorithm.

\subsection{Problem Setting}

We consider a multi-period sequential decision-making problem under uncertainty over multiple periods.
The uncertain data $\xi_1,\ldots,\xi_T$ is gradually available over $T$ periods.
We can represent these data sequences as a stochastic process denoted by $\xi_{[T]}=(\xi_1,\ldots,\xi_T)$, where $\xi_1$ is deterministic.
Then, a decision process $x_{[T]}=(x_1,\ldots,x_T)$, where $x_t\in\mathbb{R}^{n_t}$, should be made in accordance with the stochastic process, as the uncertain data $\xi_1,\ldots,\xi_T$ gradually emerges over time in $T$ periods.
It is assumed that each uncertain data $\xi_t$ has a finite moment, and the decision variable $x_{t}$ is determined based solely on the information available at each stage $t$.
The decision process has the following sequence \cite{shapiro2021lectures}:
\begin{gather*}
    x_1\leadsto \xi_2 \leadsto x_2 \leadsto \ldots \leadsto \xi_T \leadsto x_T.
\end{gather*}
Under the two processes, a $T$-stage stochastic program can be formulated in nested form as follows:
\begin{align}\label{eqn1}
    &\min_{x_1\in \mathcal{X}_1} f_1(x_1)+\mathbb{E}[\min_{x_2\in \mathcal{X}_2(x_1, \xi_2)} f_2(x_2, \xi_2)\\ 
    &+\mathbb{E}_{\cdot|\xi_{[2]}}[\cdots+\mathbb{E}_{\cdot|\xi_{[T-1]}}[\min_{x_T\in \mathcal{X}_T(x_{T-1}, \xi_T)} f_T(x_T, \xi_T)]]], \notag
\end{align}
where $\mathbb{E}_{\cdot|\xi_{[t]}}$ is the conditional expectation with respect to $\xi_{[t]}$ and $f_t(x_t,\xi_t)$ is a convex objective function in $x_t$ and dependent on $\xi_t$.
The feasible region $\mathcal{X}_t(x_{t-1}, \xi_t)$ for $x_t$, given $x_{t-1}$ and $\xi_t$, can generally be described as the intersection of sublevel sets of convex functions and hyperplanes.
Specifically, it can be expressed as $\mathcal{X}_t(x_{t-1}, \xi_t):=\{x_t:g_{t, i}(x_t, \xi_t)\leq -h_{t, i}(x_{t-1}, \xi_t), i=1,\ldots,p_t\} \cap \{x_t:l_{t, j}(x_t, \xi_t)= b_{t, j}(x_{t-1}, \xi_t), j=1,\ldots,q_t\}$.
Here, $g_{t, i}$ and $h_{t, i}$ are twice-differentiable functions in $x_t$ and $x_{t-1}$ ,respectively, while $l_{t, j}$ and $b_{t, j}$ are linear functions in $x_t$ and $x_{t-1}$ ,respectively.
The parameters $p_t$ and $q_t$ represent the number of sublevel sets of convex functions and hyperplanes, respectively.
In this study, we focus solely on the case where the feasible region is linear, and it is defined as follows.

For $t=1$,
\begin{align}
    &\mathcal{X}_1:=\{x_1:A_1(\xi_1)x_1=b_1, x_1 \geq 0\}. \notag
\end{align}
For $t=2,\ldots,T$,
\begin{align}\label{eqn2}
    \mathcal{X}_t&(x_{t-1}, \xi_t) \notag \\
    := &\{x_t: A_t(\xi_t)x_t+B_t(\xi_t)x_{t-1}=b_t(\xi_t), x_t \geq 0\}.
\end{align}
The usual MSP approach solves (\ref{eqn1}) by constructing a scenario tree that approximates the stochastic process $\xi_{[T]}$ with a finite number of realizations.
This is followed by solving a large deterministic equivalent convex optimization problem under the realized scenario tree.
However, with an increasing number of stages and nodes per stage, the scenario count grows exponentially, leading to computational intractability.
Hence, we solve the problem by decomposing it into subproblems in a stagewise manner.

In general, stagewise decomposition approaches assume that:
\begin{enumerate}[label=(A\arabic*)]
    \item\label{itm1} Stagewise independence: For stage $t=2,\ldots,T$, $\xi_t$ is independent of $\xi_{[t-1]}$.
    \item\label{itm2} Relatively complete recourse: For stage $t=2,\ldots,T$, the stage $t$ subproblem is feasible for any previous stage solution $x_{t-1}$, and for any possible realization of random observation $\xi_t$ almost surely.
\end{enumerate}
Under \ref{itm1}, we can derive the following Bellman equation by deducing from the nested structure of (\ref{eqn1}).

For $t=T,\ldots,2$,
\begin{align}\label{eqn3}
    & \mathcal{Q}_t(x_{t-1}, \xi_t)=\inf_{x_t \in \mathcal{X}_t(x_{t-1}, \xi_t)} \{f_t(x_t, \xi_t)+Q_{t+1}(x_t)\}, \notag \\
    & Q_{t+1}(x_t):= \mathbb{E}[\mathcal{Q}_{t+1}(x_t, \xi_{t+1})]
\end{align}
with $Q_{T+1}$ $\equiv$ 0.

Then, the problem (\ref{eqn1}) can be reformulated in the context of dynamic programming under the equation (\ref{eqn3}) as outlined below.

For $t=1$,
\[
    \min_{x_1\in \mathcal{X}_1} f_1(x_1)+Q_2(x_1).
\]
For $t=2,\ldots,T$,
\begin{gather}\label{eqn4}
    \min_{x_t\in \mathcal{X}_t(x_{t-1}, \xi_t)} f_t(x_t, \xi_t)+Q_{t+1}(x_t).
\end{gather}
Here, the value function $Q_t$ is demonstrated to be a convex function (refer to Appendix \ref{appendix:A} for the proof).

\subsection{Stochastic Dual Dynamic Programming} \label{section2.2}
Stochastic Dual Dynamic Programming (SDDP) is a state-of-the-art stagewise decomposition algorithm, introduced by \citet{pereira1991multi}, that is widely used to solve large-scale MSP problems.
SDDP aims to solve (\ref{eqn4}) by approximating the value function, $Q_t$, as a piecewise linear convex function based on Benders decomposition.
To enhance the piecewise linear convex function gradually, the algorithm iteratively incorporates subgradient cutting planes (referred to as cuts) while progressing.
Each iteration involves a forward and backward step, which collectively contribute to the algorithm's iterative improvement.
\citet{shapiro2011analysis} provides an analysis of the SDDP algorithm.

During the forward steps, the algorithm proceeds sequentially from $t=1$ to $T$.
For each stage $t$, we obtain the current optimal solution $x_t^*$ for the $n$-th iteration by solving the following stagewise subproblems in order, given a specific sample $\xi_t^s$:
\[
x_t^*=\underset{x_t \in \mathcal{X}_{t}(x_{t-1}, \xi_t^s)}{\arg\min} f_t(x_t, \xi_t^s)+Q_{t+1}^n(x_t).
\]
Here, $Q_{t+1}^n$ represents the approximation of the value function at the $n$-th iteration, which is expressed as $\max_{k=1,\ldots,n}\{(\beta_{t+1}^k)^\top x_t+\alpha_{t+1}^k\}$.
This value function approximation depends on the cuts constructed during the backward steps.
Then, we can reformulate the subproblem without value function, resulting in the following formulation:
\begin{align}
    &\min_{x_t \in \mathcal{X}_t(x_{t-1},\xi_t^s), \theta_{t+1}} \quad f_t(x_t, \xi_t^s)+\theta_{t+1}, \notag \\ 
    \text{s.t} \quad &\theta_{t+1} \geq (\beta_{t+1}^k)^\top x_t+\alpha_{t+1}^k, \quad k=1,\ldots,n. \notag
\end{align}
In the above program, the constraints for $\theta_{t+1}$ are referred to as cuts.
In practical implementation, it is common to introduce a trivial cut such as $\theta \geq 0$ in the initial iteration.
This approach prevents the problem from becoming unbounded when $\theta$ is selected as negative infinity in the absence of a trivial cut, and it also offers basic guidance for approximating the value function.
In the forward steps, we calculate an estimator for the optimal value, which is used as the upper bound.

In the backward steps, starting from $t=T$, for the $n$-th iteration, given sample $\xi_t^s$, the approximated value function $Q_{t}^{n}$ is updated to $Q_{t}^{n}$ by constructing cuts derived from the optimal primal-dual triple $(x_t^*, u_t^*, v_t^*)$ as follows:
\[
Q_{t}^{n+1}(x_{t-1})=\max\{Q_t^n(x_{t-1}), (\beta_t^{n+1})^\top x_{t-1}+\alpha_t^{n+1}\}.
\]
This updated value function provides a lower bound for the optimal value.
Upon the satisfaction of the stopping criterion for the upper and lower bounds, the algorithm terminates.
For more details, refer to Appendix \ref{appendix:B}.

\subsection{Improving the efficiency of SDDP}
Recent studies have focused on improving the efficiency of the SDDP algorithm.
We summarize these studies by classifying them into three main categories.

\subsubsection{Parametric Value Function Approximation}
\citet{LEE2022} introduced an algorithm called value function gradient learning (VFGL) algorithm specially designed for addressing large-scale MSP problems.
In contrast to the SDDP, which incorporates cuts to approximate the value function iteratively, VFGL utilizes a specific parametric convex function, denoted as $\hat{Q_t}(x_{t-1}, \theta_t)$, to approximate the true value function $Q_t(x_{t-1})$.
By learning the parameters $\theta_t$, VFGL tackles large-scale MSP problems by optimizing these parameters.
The algorithm utilizes stochastic gradient descent-based optimization to minimize the discrepancy between the gradients of the parametric function and that of the true value function, obtained through the duality of optimization problems \cite{boyd2004convex}.

\subsubsection{Selection of subgradient cutting planes}
There has been research that focused on selecting a subset of generated cuts based on heuristic conditions.
The territory algorithm proposed by \citet{pfeiffer2012two} identifies a subset of cuts using specific criteria, and the test of usefulness subsequently eliminates redundant cuts from this subset.
However, this approach can be computationally expensive and challenging to implement in practice.
\citet{de2015improving} proposed two cut selection strategies: the last-cuts strategy, which conducts a piecewise linear lower bound by selecting recent cuts, and the Level N dominance strategy, which conducts a piecewise lower bound by selecting cuts that have been activated at least $n$ times for the trial solution.
The experimental results showed that the Level 1 dominance strategy outperforms other strategies.

\subsubsection{Generation of subgradient cutting planes}
\citet{dai2021neural} proposed an algorithm called neural stochastic dual dynamic programming ($\nu$-SDDP), which employs a neural network to generate cuts for approximating the value function.
This approach utilizes the problem context vector that embeds information about the MSP problem.
Through meta-learning, the context vector is mapped to the corresponding cut information, represented by $\alpha_t$ and $\beta_t$.
The elements of the context vector, which are probabilities or functions associated with the stochastic elements in the problem, are sampled from specific probability distributions.
This facilitates the derivation of solutions for a family of problems, rather than being limited to a single instance.
The study demonstrated that $\nu$-SDDP offers significant improvements in the computational efficiency of SDDP.

\subsection{Transformer}
The Transformer model, proposed by \citet{vaswani2017attention}, presents a revolutionized encoder-decoder framework by introducing an all-attention architecture without recurrent neural networks (RNNs) and convolutional neural networks (CNNs). 
Unlike existing sequence models, the Transformer leverages attention mechanisms, including self-attention, to calculate attention scores that determine the relevance between target and source elements \cite{bahdanau2014neural, luong2015effective, kim2017structured}.
This approach allows the model to focus on relevant score elements during processing.
The Transformer introduced the key concepts: self-attention and multi-head attention.
Self-attention mechanism enables the model to achieve computational advantages, efficient modeling of long-range dependencies, and improved interpretability.
Multi-head attention allows the model to gather attention scores and information from different perspectives and to offer computational benefits.
Consequently, the Transformer is able to effectively handle long-term dependencies and non-parallelization issues that plagued the existing sequence models.
\section{Model}
The $\nu$-SDDP algorithm, while enhancing the efficiency of SDDP, still exhibits certain limitations.
Unlike the SDDP algorithm, which relies on previous cuts for cut generation, $\nu$-SDDP exclusively employs the problem context vector for cut generation, potentially failing to fully capture the relationships among cuts.
Moreover, $\nu$-SDDP is restricted to linear programs and imposes a predetermined limit on the number of cuts.
To address these limitations, we propose a novel approach called TranSDDP.
TranSDDP takes advantage of a Transformer-based generative model to generate cuts for MSP problems defined by a parametric family of stochastic elements $\xi_t$ and approximates the value function by utilizing these generated cuts.

\subsection{Input and Output Sequence}
We represent the input sequence as a set of parameters that define the probability distribution for the stochastic elements in $A_t$, $B_t$, and $b_t$ of the feasible region (\ref{eqn2}).
The input sequence, denoted as 
\begin{gather*}
    \{(\lambda_{A_{t, i}}^1,\ldots,\lambda_{A_{t, i}}^{M_{A_{t, i}}}, \lambda_{B_{t, i}}^1, \ldots,\lambda_{B_{t, i}}^{M_{B_{t, i}}},\\
    \lambda_{b_{t, i}}^1,\ldots,\lambda_{b_{t, i}}^{M_{b_{t, i}}}, \tilde{t})_{i=1}^N\},
\end{gather*}
captures the parameters associated with the probability distribution for each stochastic element, where $\lambda$ represents these parameters.
We define such input sequences for each stage $t$ value, ranging from $1$ to $T-1$.
The subscript of each element represents the coefficient associated with the $i$-th constraint in the feasible region at stage $t$, while the superscript represents the $j$-th parameter among $M_c$ parameters that define the distribution of the corresponding coefficient $c$.
We assume that the parameter $\lambda$ is sampled from a prior distribution to consider a parametric family.
That is, $c \sim p_c(\cdot|\lambda_c^1,\ldots,\lambda_c^{M_c})$ and $\lambda_c^j \sim p_{\lambda_c^j}(\cdot)$ for $j=1,\ldots,M_c$.
By formulating the input in this manner, even if perturbations occur in the problem, it becomes feasible to promptly obtain a solution through the learned model without re-solving.
The variable $N$ denotes the number of constraints, and $\tilde{t}$ represents the relative position for stage $t$, defined as $t/(T-1)$.

TranSDDP generates a set of parameters representing cuts and category indicator that denotes the start, middle, and end of the generated sequence.
This output is represented as:
\[
\{\Tilde{\beta}_k, \Tilde{\alpha}_k, \Tilde{\tau}_k\}_{k=1}^K,
\]
where $\Tilde{\beta}_k \in \mathbb{R}^d$ and $\Tilde{\alpha}_k \in \mathbb{R}$ denote the gradient of each decision variable and the intersection of the $k$-th cut, respectively.
Here, $d$ refers to the dimension of the decision variables.
The category information, referred to as a token, is encoded using a one-hot vector denoted by $\tau_k$.
This vector indicates the position of the output in the sequence, such that $\tau_k=(1, 0, 0, 0), (0, 1, 0, 0), (0, 0, 1, 0)$, and $(0, 0, 0, 1)$ represents the padding, start, middle, and end position, respectively.
The TranSDDP model generates a sequence of $\Tilde{\beta}_k$ and $\Tilde{\alpha}_k$ that construct a piecewise linear convex function until the token indicating the end of the output sequence is encountered.
It is important to note that the value $K$ is not predetermined.
The approximated value function $\Tilde{Q}_t^K(x_{t-1})$ is given by $\max_{k=1,\ldots,K} \{(\Tilde{\beta}_k)^\top x_{t-1}+\Tilde{\alpha}_k\}$.

\subsection{Model Architecture}
TranSDDP is a modified version of the Transformer architecture specifically designed to address the unique requirements of the problem at hand.
It incorporates several modifications to handle the continuous nature of the input and output sequences, excluding $\tau_k$.
These modifications include replacing the input and output embedding layers with linear layers and adjusting the softmax layer of the Transformer to exclusively apply to vectors corresponding to $\tau_k$.
To further enhance performance, TranSDDP incorporates positional encoding solely in the output sequence, which serves as the input for the decoder, as the positional information of the input sequence is deemed less crucial.

It is worth noting that the size of the input sequence is fixed, and the interrelationships between elements are relatively insignificant.
Consequently, applying a self-attention layer to the input sequence may lead to increase computational complexity without providing significant performance improvements.
Hence, we propose a variant of the TranSDDP model called TranSDDP-Decoder, which exclusively utilizes the decoder component of TranSDDP as an alternative solution.
Details of the architecture for both TranSDDP and TranSDDP-Decoder can be found in Appendix \ref{appendix:C}.

\subsection{Dataset} \label{section3.3}
To construct the dataset, we apply the SDDP algorithm to solve the problem defined in this study.
The dataset can be represented as $\mathcal{D}_S := \{z_s:=(\Lambda, \Tilde{t}, \{\beta_k, \alpha_k, \tau_k\}_{k=1}^K)_s\}_{s=1}^S$.
Here, $\Lambda$ refers to the sampled parameters that capture the probability distribution for the stochastic elements $(A_t, B_t, b_t)$ used to define the problem.
Specifically, $\Lambda$ is composed of individual parameters $\lambda^j$ drawn from the probability distribution $p_{\lambda^j}(\cdot)$, with $j$ ranging from $1$ to $M$, i.e., $\Lambda = \{\lambda^j\}_{j=1}^M$ where $ \lambda^j \sim p_{\lambda^j}(\cdot)$. 

Since the SDDP algorithm generates cuts iteratively until a convergence criterion is met, it is possible to obtain data with an unusually large number of cuts.
To address this, we utilize a dataset in which the number of cuts falls below the $100(1-\alpha)$ percentile, with $\alpha$ set to $0.025$.
Specifically, we identify data points with cut counts exceeding the top 2.5\% as outliers and exclude them from the dataset.
Afterward, we divide the resulting dataset into six folds with a ratio of 5:1 for training and validation data and perform cross-validation.
In the case of the energy planning problem, the dataset comprises 17,652 data points, and instances with 80 or more cuts are considered outliers.
For the financial planning problem, the dataset consists of 23,580 data points, and instances with 40 or more cuts are identified as outliers.
For the production planning problem, the dataset contains 19,186 data points, and instances with 65 or more cuts are regarded as outliers.
We evaluate the test dataset, which consists of 100 data points for each problem.

\subsection{Learning System} \label{section3.4}
Based on the dataset generated through the methodology outlined in section \ref{section3.3}, the model parameters $W$ are optimized by minimizing a loss function that combines the mean squared error (MSE) and cross-entropy (CE) loss.
This loss function quantifies the discrepancy between the target output sequence $\{\beta_k, \alpha_k, \tau_k\}$ and the predicted output sequence $\{\Tilde{\beta}_k, \Tilde{\alpha}_k, \Tilde{\tau}_k\}$.
The loss function is defined as follows:
\begin{align*}
    L(W):= \frac{1}{N} \frac{1}{K} \sum_{i=1}^N \sum_{k=2}^K &\bigg[||\beta_k^{(i)} - \Tilde{\beta}_k^{(i)}||_2^2 + (\alpha_k^{(i)} - \Tilde{\alpha}_k^{(i)})^2 \\
    & - \sum_{c=1}^4 \tau_{k, c}^{(i)} log(\Tilde{\tau}_{k, c}^{(i)})\bigg],
\end{align*}
where $N$ denotes the batch size.

The encoder of TranSDDP takes the input sequence, and the decoder of TranSDDP takes the target output sequence ranging from $k=1$ to $K-1$.
TranSDDP then generates the sequence of predicted cuts and category tokens for $k$ ranging from $k=2$ to $K$.
To initiate the process, the decoder requires a designated starting token as input.
To fulfill the prerequisite, we substitute the starting token with the trivial cut described in Section \ref{section2.2} and employ this trivial cut as the initial cut corresponding to $k=1$, indicating the first element in the sequence.
The teacher forcing technique is employed by feeding the decoder with the target sequence instead of the predicted sequence.
The training procedure is presented in Algorithm \ref{alg:TranSDDP}, and the design of the learning system, including validation and test procedures, is illustrated in Appendix \ref{appendix:D}.

\begin{algorithm}
\caption{TranSDDP}\label{alg:TranSDDP}
\begin{algorithmic}
    \STATE \textbf{Initialize:} $\mathcal{D}_0$ (dataset), $M$ (number of epochs), $\mathcal{B}$ (number of batch), $(\beta_1, \alpha_1, 0, 1, 0, 0)$ (initial cut and token), $m, v \gets 0$
    \FOR[Creating dataset]{$s=1,\ldots,S$}
        \STATE Sample a stochastic element's distribution parameters $\Lambda = \{\lambda^j\}_{j=1}^M \sim p_{\lambda^j}(\cdot)$
        \STATE $\{\beta_k, \alpha_k, \tau_k\}_{k=2}^K = \textit{SDDP}(\Lambda)$
        \STATE Update the dataset \\ 
        $\mathcal{D}_s=\mathcal{D}_{s-1} \cup (\Lambda, \Tilde{t}, \{\beta_k, \alpha_k, \tau_k\}_{k=1}^K)$
    \ENDFOR
    \FOR[Training the model]{epoch = $1,\ldots,M$}
        \FOR{batch = $1,\ldots,\mathcal{B}$}
            \STATE Sample $z_l \sim D_S$
            \STATE $\{\Tilde{\beta}_k, \Tilde{\alpha}_k, \Tilde{\tau}_k\}_{k=2}^K = \textit{TranSDDP}(z_l)$
            \STATE Update parameters $W$ using the Adam optimizer:
            \begin{ALC@g}
                \STATE \quad $m \gets \gamma_1 m + (1 - \gamma_1) {\nabla}_W L(W)$
                \STATE \quad $v \gets \gamma_2 v + (1 - \gamma_2) ({\nabla}_W L(W) \odot {\nabla}_W L(W))$
                \STATE \quad $\hat{m} \gets \frac{m}{1-\gamma_1}, \hat{v} \gets \frac{v}{1-\gamma_2}$
                \STATE \quad $w \gets w - \epsilon \frac{\hat{m}}{\sqrt{\hat{v}}+\delta}$
            \end{ALC@g}
            
        \ENDFOR
    \ENDFOR

\end{algorithmic}
\end{algorithm}
\section{Experiments} \label{section4}
In this section, we provide three examples to demonstrate the effectiveness of TranSDDP in various applications, namely energy planning, financial planning, and production planning.
The problem instances are cast as multistage convex stochastic programs.
These problem instances have been extensively investigated and analyzed within the community of optimization.
It is worth noting that these problems can be easily adapted and further explored.

In the numerical evaluation, we examine the performances of various methods, including MSP, SDDP, Level 1 dominance (L1), VFGL, $\nu$-SDDP with 40 predetermined numbers of cuts, TranSDDP, and TranSDDP-Decoder, on 7- and 10-stage multistage stochastic convex problems with 5 and 3 scenario branches, respectively.
Table \ref{tab1} provides information on the number of variables and constraints for each problem when represented as a scenario tree.
\begin{table}[t]
\caption{Variable and constraint counts of numerical experiments}
\label{tab1}
\vskip 0.1in
\centering
\resizebox{0.48\textwidth}{!}{
  \begin{tabular}{llll}
  \toprule
    Stage & Problem & \# of Variables & \# of Constraints \\
    \midrule
    \multirow{3}{*}{$T$ = 7} & Enerrgy Planning & 78,124 & 136,717 \\
    & Financial Planning & 46,873 & 54,684 \\
    & Production Planning & 128,904 & 121,092 \\
    \midrule
    \multirow{3}{*}{$T$ = 10} & Energy Planning & 118,096 & 206,668 \\
    & Financial Planning & 78,729 & 98,410 \\
    & Production Planning & 206,667 & 186,985 \\
    \bottomrule
  \end{tabular}
  }
  \vspace{-0.6cm}
\end{table}

We evaluate the performance of the algorithms by analyzing both the solution quality obtained and the total computational time, which encompasses both the evaluation time and the training time.
In this context, the evaluation time refers to the duration required to solve the problem.
The solution quality of the algorithms is evaluated by measuring its error ratio, which quantifies the difference between the algorithm's objective value and that of MSP.
MSP is utilized as a reference standard for comparative analysis due to its ability to locate the global optimum within a finite tree of scenarios.
The error ratio for the candidate methods is calculated using the following formula:
\[
error\,ratio = \frac{|obj(\textbf{candidate}) - obj(\textbf{MSP})|} {|obj(\textbf{MSP})|}, 
\]
where $obj$ means the objective value.
Based on 100 repeated test experiments, the error ratio is presented as the mean value with standard errors. 
Similarly, the evaluation time is reported as the mean value.
We also conduct a feasibility test by computing the infeasibility ratio, which represents the proportion of infeasible problems out of the total number of problems.
By incorporating the cuts generated by the model trained up to the corresponding epoch as additional constraints in the original problem, we verify the feasibility of the problems obtained with the TranSDDP and TranSDDP-Decoder models.
Furthermore, to validate the accuracy of the cut approximation, we compare the value function with its approximations from the SDDP, TranSDDP, and TranSDDP-Decoder algorithms.
A detailed explanation of this process can be found in Appendix \ref{appendix:F}.
All experiments were performed using an AMD Ryzen 5 5600X processor with 48 GB of RAM and GeForce RTX 3060.
And each subproblem was solved using the CVXPY 1.1.17 library and MOSEK 9.3.11 solver.

\subsection{Energy Planning}
We investigate an energy planning (EP) problem that involves determining the optimal electricity generation levels for hydro and thermal power plants.
The objective is to satisfy a predetermined demand while minimizing the sum of expected production costs and penalties associated with reservoir levels.
This example is a simplified version of a hydroelectric system discussed in \citet{guigues2014sddp}.
In this system, the hydro plant has lower production costs but is constrained by the reservoir level, and the water inflow to the reservoir is stochastic.
The stochastic water inflow at each stage $t$ is modeled by a normal distribution: $I_t \sim \mathcal{N}(\mu_I, {\sigma_I}^2)$, where $\mu_I \sim \mathcal{U}(15, 25), \sigma_I \sim \mathcal{U}(4, 6)$. 
The formulation and details of the decision variables and parameters can be found in Appendix \ref{appendix:E.1}.

\begin{figure*}[t]
    \centering
    \subfigure[Energy Planning]{
    \includegraphics[width=0.653\columnwidth]{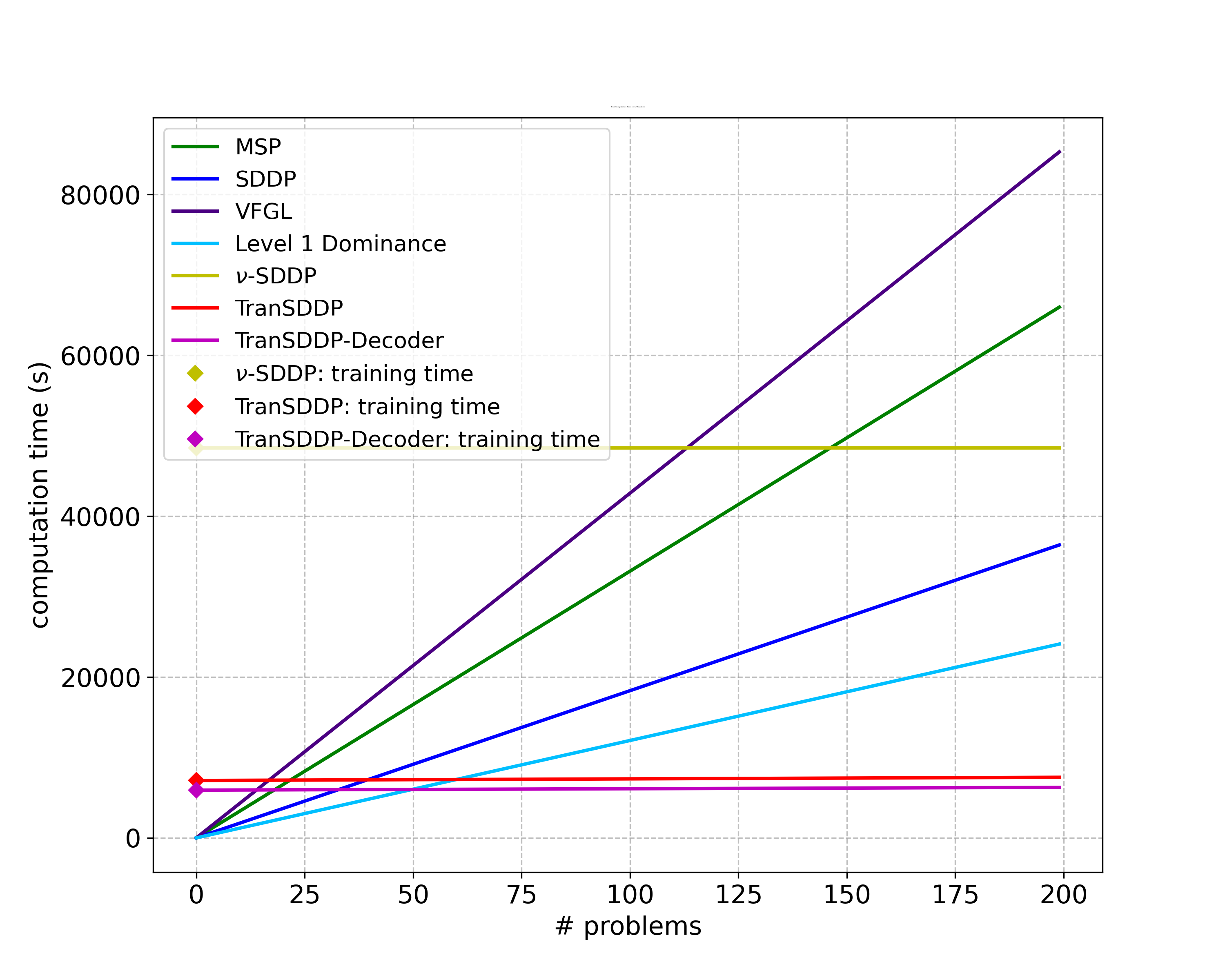}
    \label{fig1(a)}
    }
    \subfigure[Financial Planning]{
    \includegraphics[width=0.653\columnwidth]{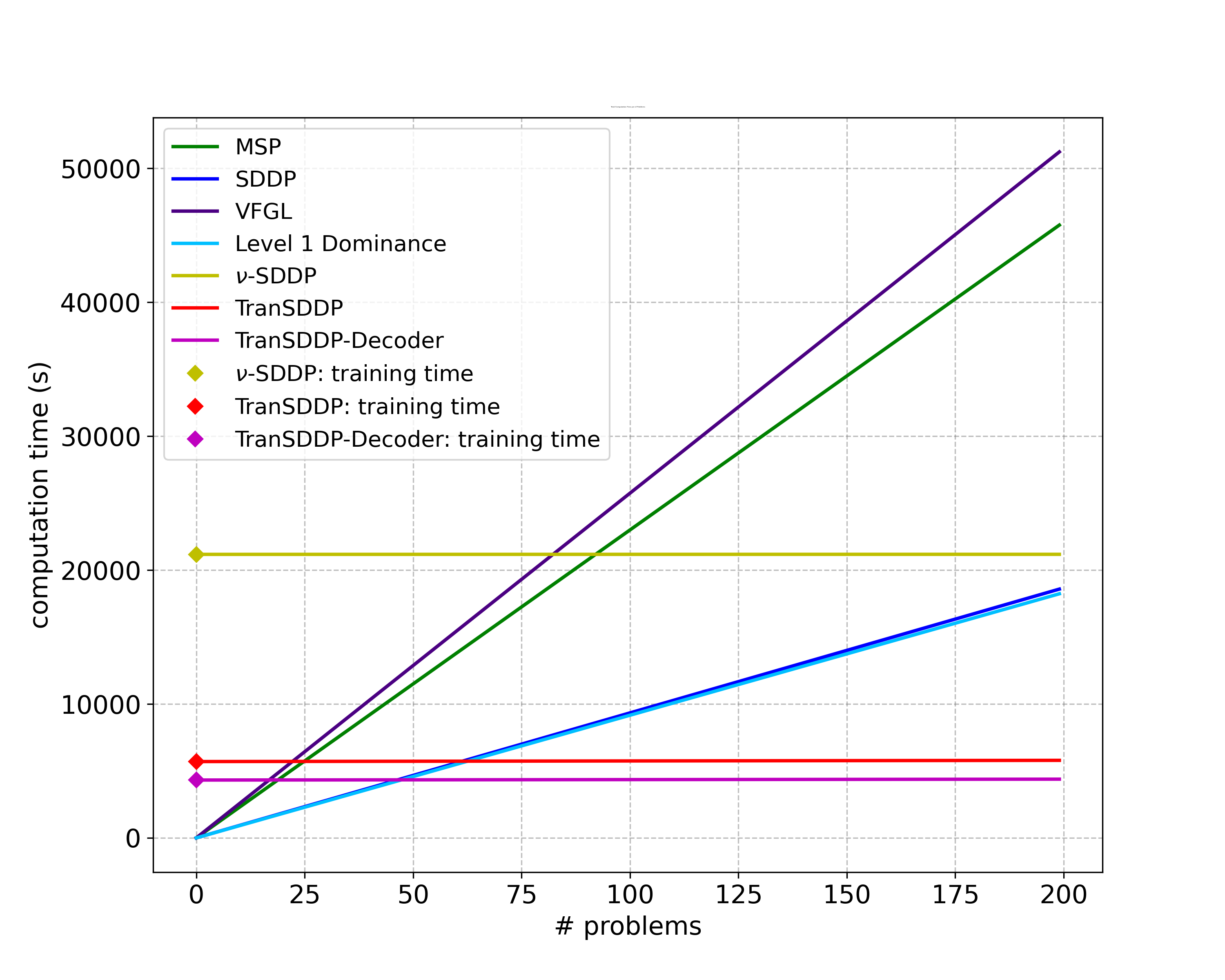}
    \label{fig1(b)}
    }
    \subfigure[Production Planning]{
    \includegraphics[width=0.653\columnwidth]{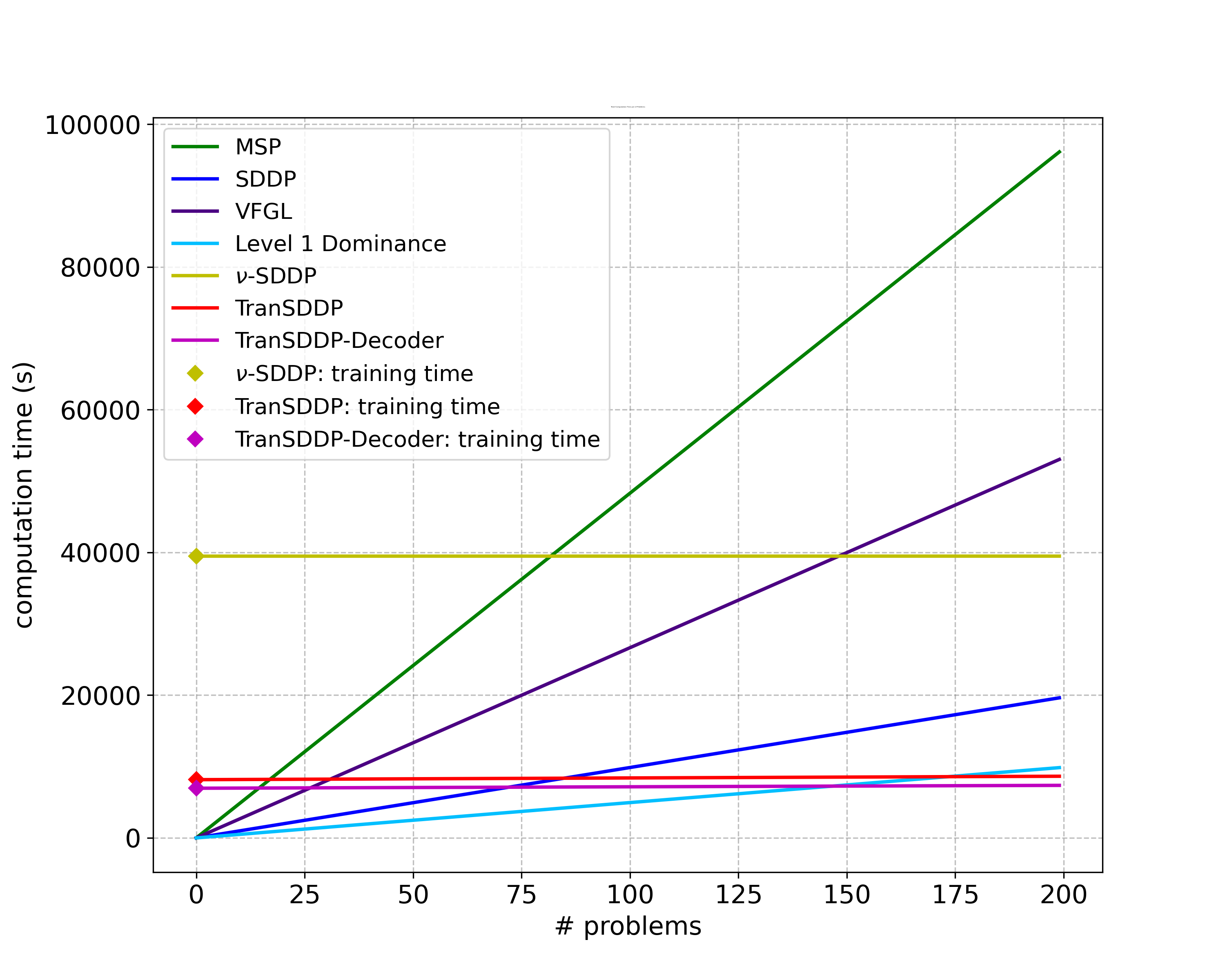}
    \label{fig1(c)}
    }
    \caption{
    Time elapsed per problem
    }
\end{figure*}

\begin{figure*}[t]
    \centering
    \subfigure[Energy Planning]{
    \includegraphics[width=0.653\columnwidth]{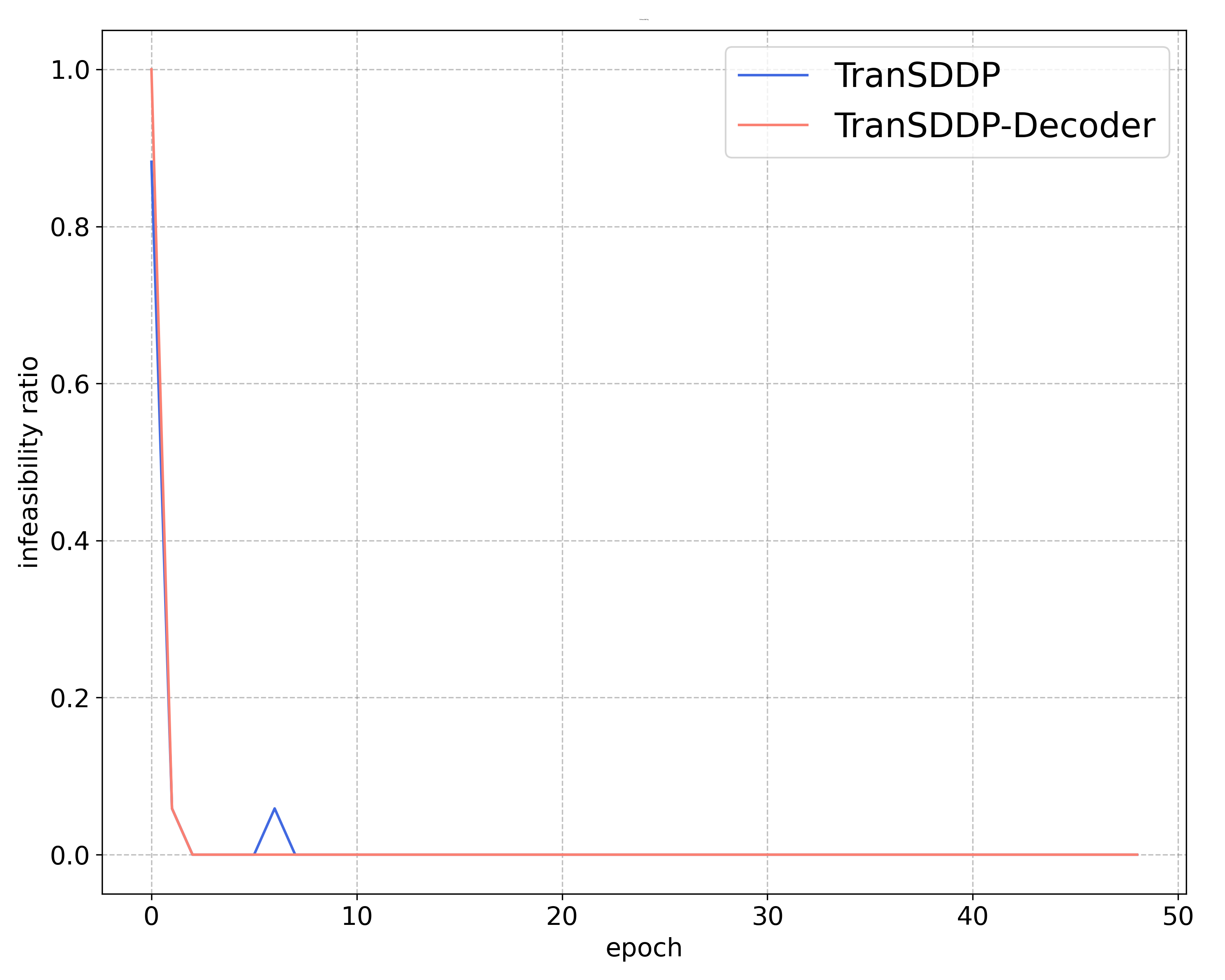}
    \label{fig2(a)}
    }
    \subfigure[Financial Planning]{
    \includegraphics[width=0.653\columnwidth]{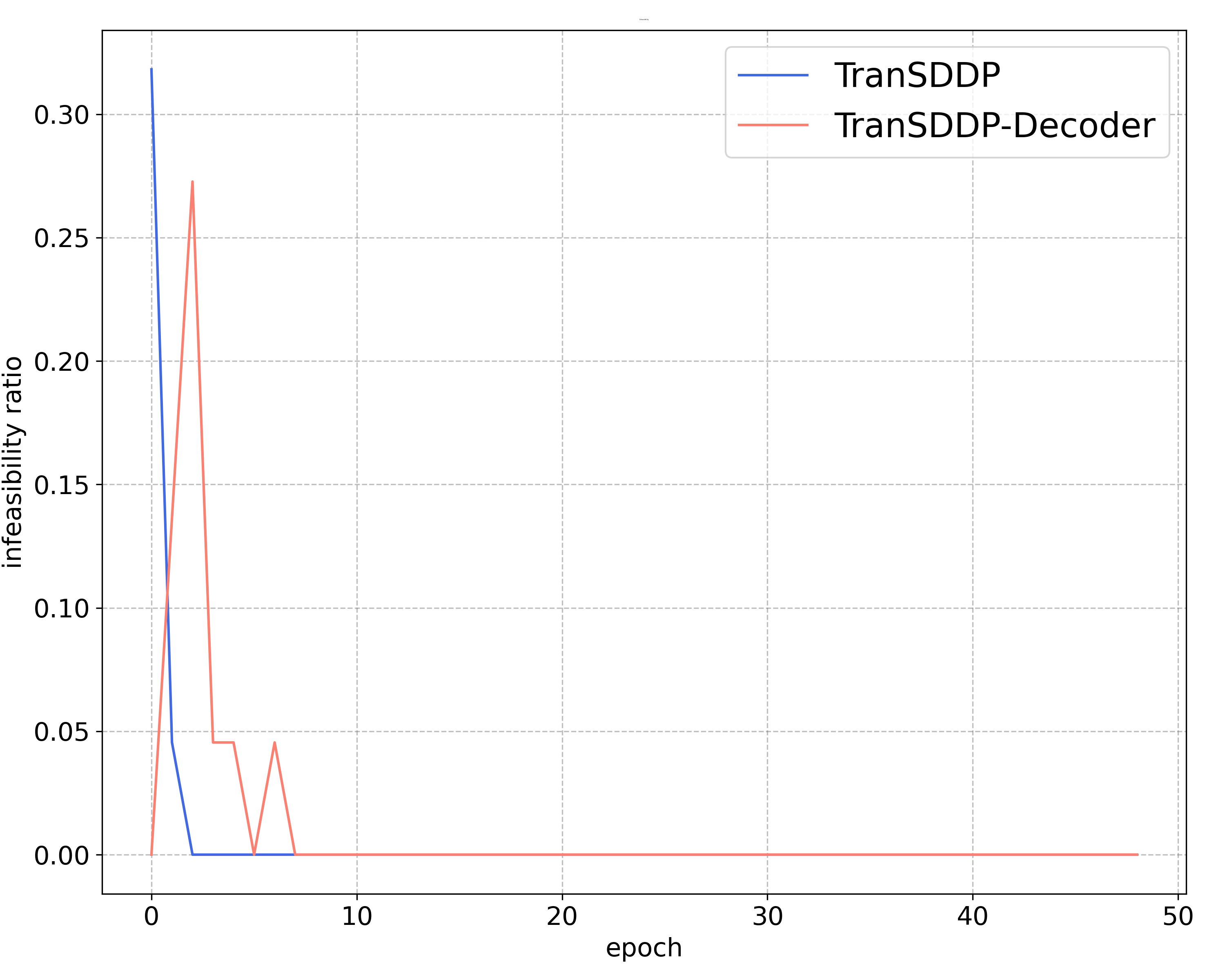}
    \label{fig2(b)}
    }
    \subfigure[Production Planning]{
    \includegraphics[width=0.653\columnwidth]{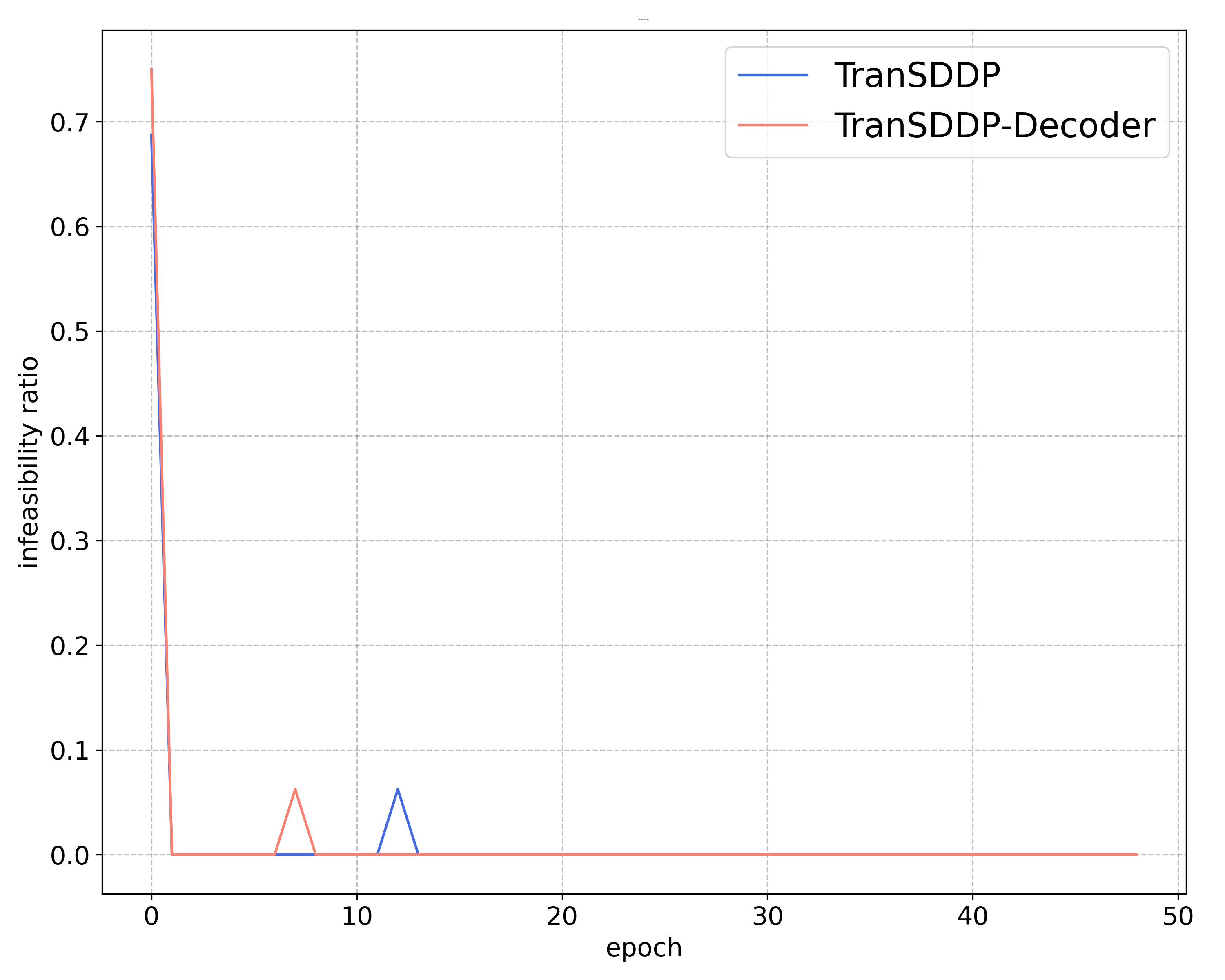}
    \label{fig2(c)}
    }
    \caption{
    Infeasibility ratio per epoch
    }
    \label{fig2}
\end{figure*}

\begin{table}[t]
\caption{Performance comparisons for EP problem \quad\quad\quad}
\label{tab2}
\vskip 0.1in
\centering
\resizebox{0.5\textwidth}{!}{
  \begin{tabular}{lllll}
  \toprule
    Task & Algorithm & Error ratio & Evaluation time (s) & Training time \\
    \midrule
    \multirow{7}{*}{$T$ = 7} & MSP & - & 331.58 & - \\
    & SDDP & 3.349 $\pm$ 2.698\% & 183.01 & - \\
    & L1 & 0.326 $\pm$ 0.379\% & 121.096 & - \\
    & VFGL & 1.169 $\pm$ 0.822\% & 428.559 & - \\
    & $\nu$-SDDP & 40.410 $\pm$ 29.742\% & 0.02 & 13h 27m \\
    & TranSDDP & 1.191 $\pm$ 0.701\% & 1.990 & 1h 59m \\
    & TranSDDP-Decoder & 1.010 $\pm$ 0.548\% & 1.712 & 1h 39m \\
    \midrule
    \multirow{7}{*}{$T$ = 10} & MSP & - & 902.73 & - \\
    & SDDP & 3.441 $\pm$ 3.357\% & 450.12 & - \\
    & L1 & 0.35 $\pm$ 0.43\% & 179.776 & - \\
    & VFGL & 1.796 $\pm$ 1.100\% & 438.237 & - \\
    & $\nu$-SDDP & 68.070 $\pm$ 5.459\% & 0.023 & 23h 22m \\
    & TranSDDP & 2.337 $\pm$ 1.736\% & 2.100 & 2h 30m \\
    & TranSDDP-Decoder & 3.826 $\pm$ 2.018\% & 1.784 & 2h 6m \\
    \bottomrule
  \end{tabular}
  }
  \vspace{-0.6cm}
\end{table}

The results summarized in Table \ref{tab2} show that our proposed models exhibit a higher error ratio compared to the L1 algorithm, comparable performance to VFGL, and an improvement over the other algorithms.
Specifically, the mean and standard error of the error ratio for our proposed models are lower than those for SDDP.
This finding supports that the TranSDDP and TranSDDP-Decoder models provide more accurate and stable approximations of the value function.
While the SDDP algorithm approximates the value function through a sampling of stochastic elements, the proposed models utilize information regarding the parameters from which stochastic elements are sampled to approximate the value function.
They minimize the mean squared error between the value function approximations and the multiple cuts generated by SDDP for the problems instantiated by those parameters.
Consequently, our models yield more robust approximations of the value function compared to SDDP.
This can be validated by examining the comparison between the value function and its approximations produced by SDDP, TranSDDP, and TranSDDP-Decoder in Appendix \ref{appendix:F.1.1}.
In the context of the EP problem, the limitations of $\nu$-SDDP become apparent as it is unable to handle the problem due to its non-linear programming formulation.

As demonstrated in Table \ref{tab2}, we observe that the evaluation time significantly increases for MSP, SDDP, and L1 algorithms as the number of stages defining the problem grows.
However, for the TranSDDP and TranSDDP-Decoder algorithms, the evaluation time remains stable and minimal.
The notable advantage of having a considerably small evaluation time, combined with the capacity to promptly generate solutions for problems defined by a parametric family using the trained model, becomes apparent when faced with a larger number of problems to solve.
In Figure \ref{fig1(a)}, we present a comparison of the total computational time required to solve a 7-stage EP problem as the number of problems to be solved increases.
TranSDDP and TranSDDP-Decoder show a nearly constant total computational time, irrespective of the number of problems.
Moreover, we demonstrate the computational superiority of the TranSDDP (TranSDDP-Decoder) algorithm over SDDP for 39 (33) or more problems, even when considering training time.

We verify the feasibility of the cuts generated by our models in the context of a 7-stage problem. 
As illustrated in Figure \ref{fig2(a)}, during the initial phases of training, infeasible cuts are generated.
However, as the training progresses, no infeasibilities are observed.
Consequently, the TranSDDP and TranSDDP-Decoder algorithms can provide solutions of satisfactory quality while enhancing computational efficiency.
Further experimental findings, including the training/validation loss/error, can be found in Appendix \ref{appendix:F.1.2}.

\subsection{Financial Planning}
Next, we explore the financial planning (FP) problem modeled as a continuous-time portfolio optimization problem by \citet{merton1969lifetime}.
The objective is to optimize the allocation of wealth between bonds and stocks, as well as strategic decisions on consumption levels, with the aim of maximizing the overall utility of consumption and final wealth.
Modifying Merton's problem slightly, the rate of return on stock investments is introduced as a stochastic element that follows a log-normal distribution.
Specifically, the rate of return is represented as $\log(r^{stock}) \sim \mathcal{N}((\mu-{\sigma^2}/2)\Delta t, \sigma^2 \Delta t)$, where $\mu \sim \mathcal{U}(0.04, 0.08)$ and $\sigma \sim \mathcal{U}(0.15, 0.25)$.
Details about the FP problem are in Appendix \ref{appendix:E.2}.

The analysis of the results is conducted in the same manner as for the previous problem.
The performance shown in Table \ref{tab3} demonstrates that our proposed models exhibit superior results compared to the SDDP and L1 algorithms, although they do not achieve the performance level of VFGL.
Similar to the previous example, the $\nu$-SDDP algorithm encounters challenges in solving non-linear convex optimization problems.

As presented in Table \ref{tab3}, we demonstrate that the evaluation time of our proposed models exhibits minimal increase as the problem size grows.
Consequently, the total computational time of our proposed models remains relatively stable even when the number of problems to be solved increases due to similar reasons as the previous example.
This is supported by the outcomes illustrated in Figure \ref{fig1(b)}.
Specifically, Figure \ref{fig1(b)} provides a visual representation of the computational advantage of the TranSDDP (TranSDDP-Decoder) algorithm over SDDP in solving 62 (47) or more 7-stage FP problems.

Figure \ref{fig2(b)} demonstrates the convergence of the proportion of infeasible problems to zero as the model is trained.
Appendix \ref{appendix:F.2} includes a comparison of the value function and its approximations, as well as results obtained from the training and validation process.

\begin{table}[t]
\caption{Performance comparisons for FP problem \quad\quad\quad}
\label{tab3}
\vskip 0.1in
\centering
\resizebox{0.5\textwidth}{!}{
  \begin{tabular}{lllll}
  \toprule
    Task & Algorithm & Error ratio & Evaluation time (s) & Training time \\
    \midrule
    \multirow{7}{*}{$T$ = 7} & MSP & - & 229.93 & - \\
    & SDDP & 1.782 $\pm$ 1.192\% & 93.35 & - \\
    & L1 & 1.283 $\pm$ 1.096\% & 91.595 & - \\
    & VFGL & 0.200 $\pm$ 0.160\% & 257.387 & - \\
    & $\nu$-SDDP & 515.722 $\pm$ 0.802\% & 0.021 & 5h 52m \\
    & TranSDDP & 0.962 $\pm$ 0.199\% & 0.460 & 1h 35m \\
    & TranSDDP-Decoder & 0.611 $\pm$ 0.198\% & 0.335 & 1h 12m \\
    \midrule
    \multirow{7}{*}{$T$ = 10} & MSP & - & 559.61 & - \\
    & SDDP & 2.848 $\pm$ 1.647\% & 144.57 & - \\
    & L1 & 1.630 $\pm$ 1.360\% & 177.941 & - \\
    & VFGL & 0.110 $\pm$ 0.082\% & 284.863 & - \\
    & $\nu$-SDDP & 317.890 $\pm$ 0.448\% & 0.024 & 11h 4m \\
    & TranSDDP & 1.704 $\pm$ 0.209\% & 0.630 & 1h 50m \\
    & TranSDDP-Decoder & 1.364 $\pm$ 0.208\% & 0.490 & 1h 17m \\
    \bottomrule
  \end{tabular}
  }
  \vspace{-0.6cm}
\end{table}

\subsection{Production Planning}
We examine a production planning (PP) problem, which involves optimizing the production quantities of three products at each stage to meet uncertain demand while minimizing costs related to manufacturing, outsourcing, and inventory holding.
This problem has been extensively researched in the literature \citep{wagner1958dynamic, shapiro1993mathematical, karimi2003capacitated}.
In the experiment, demand of product $i$ at stage $t$ are generated from a normal distribution: $d_{i, t} \sim \mathcal{N}(\mu_{d_i}, {\sigma_{d_i}}^2)$, where $\mu_{d_1} \sim \mathcal{U}(3, 6), \sigma_{d_1} \sim \mathcal{U}(0.2, 0.4), \mu_{d_2} \sim \mathcal{U}(1.5, 4), \sigma_{d_2} \sim \mathcal{U}(0.1, 0.2), \mu_{d_3} \sim \mathcal{U}(1, 2)$, and $\sigma_{d_3} \sim \mathcal{U}(0.05, 0.1)$.
Details of this problem can be found in Appendix \ref{appendix:E.3}.

The results in Table \ref{tab4} indicate that our proposed models exhibit relatively lower accuracy compared to other algorithms, but demonstrate a slight improvement compared to the $\nu$-SDDP algorithm.
This result confirms that the utilization of information from previous cuts in the generation of new cuts, which is a distinctive point of our model in contrast to the $\nu$-SDDP algorithm, is beneficial.
Furthermore, we demonstrate that both the TranSDDP and TranSDDP-Decoder algorithms show notable computational advantages over the benchmark algorithms, similar to the findings observed in the preceding experiments.
Figure \ref{fig1(c)} illustrates that the TranSDDP (TranSDDP-Decoder) algorithm has a computational benefit over SDDP when solving 85 (72) or more 7-stage PP problems.
Additionally, Figure \ref{fig2(c)} verifies that our proposed models do not generate infeasible problems.
Further experimental results for the current experiment, resembling those presented earlier, are available in Appendix \ref{appendix:F.3}.

\begin{table}[t]
\caption{Performance comparisons for PP problem \quad\quad\quad}
\label{tab4}
\vskip 0.1in
\centering
\resizebox{0.5\textwidth}{!}{
  \begin{tabular}{lllll}
  \toprule
    Task & Algorithm & Error ratio & Evaluation time (s) & Training time \\
    \midrule
    \multirow{7}{*}{$T$ = 7} & MSP & - & 486.046 & - \\
    & SDDP & 0.072 $\pm$ 0.115\% & 98.637 & - \\
    & L1 & 0.239 $\pm$ 0.762\% & 49.486 & - \\
    & VFGL & 0.692 $\pm$ 0.687\% & 266.514 & - \\
    & $\nu$-SDDP & 7.112 $\pm$ 2.648\% & 0.017 & 10h 58m \\
    & TranSDDP & 3.628 $\pm$ 3.341\% & 2.410 & 2h 16m \\
    & TranSDDP-Decoder & 0.838 $\pm$ 0.831\% & 2.018 & 1h 56m \\
    \midrule
    \multirow{7}{*}{$T$ = 10} & MSP & - & 1103.122 & - \\
    & SDDP & 0.076 $\pm$ 0.110\% & 151.908 & - \\
    & L1 & 0.096 $\pm$ 0.259\% & 78.630 & - \\
    & VFGL & 0.440 $\pm$ 0.430\% & 282.234 & - \\
    & $\nu$-SDDP & 2.770 $\pm$ 2.030\% & 0.02 & 19h 3m \\
    & TranSDDP & 3.580 $\pm$ 3.510\% & 2.594 & 2h 18m \\
    & TranSDDP-Decoder & 0.967 $\pm$ 0.182\% & 2.235 & 1h 57m \\
    \bottomrule
  \end{tabular}
  }
  \vspace{-0.6cm}
\end{table}
\section{Conclusion}
We propose novel models, TranSDDP and TranSDDP-Decoder, which utilize Transformer architecture for stagewise decomposition in large-scale multistage stochastic optimization problems. 
These algorithms exploit the sequential nature of the problem to improve performance.
The proposed TranSDDP and TranSDDP-Decoder models outperform benchmark algorithms in terms of computational time while also achieving a high level of accuracy in solving multistage stochastic convex problems.
Numerical experiments conducted on energy planning, financial planning, and production planning problems demonstrate the effectiveness of the TranSDDP and TranSDDP-Decoder models in solving real-world multistage stochastic convex problems with improved computational time.
Specifically, in contrast to conventional algorithms that require solving the problem anew in the presence of perturbations, these algorithms offer advantages when a significant number of similar problems with slight variations need to be solved within tight time constraints.
Although the proposed models show promising results, there is still room for improvement in terms of dataset collection.
Incorporating transfer learning is expected to effectively address these limitations.

\section*{Acknowledgements}
This research was supported by the National Research Foundation of Korea (NRF) grant funded by the Ministry of Science and ICT (NRF-2022M3J6A1063021 and RS-2023-00208980).


\bibliography{main}
\bibliographystyle{icml2023}

\newpage
\appendix
\onecolumn
\section{Proof of convexity of value function} \label{appendix:A}
Consider the stage $t$ value function $Q_{t+1}(x_t,\xi_{[t]})=\mathbb{E}_{\cdot|\xi_{[t]}}[{\mathcal{Q}}_{t+1}(x_{t},\xi_{[t+1]})]$. 
Its convexity with respect to $x_t$ can be shown by the convexity of $\mathcal{Q}_{t+1}(x_t,\xi_{[t+1]})$, because expectation preserves convexity. 
We show the convexity of $\mathcal{Q}_{t+1}(x_t,\xi_{[t+1]})$ by mathematical induction. 

Assume that $\mathcal{Q}_{t+1}(x_t,\xi_{[t+1]})$ is convex in $x_t$. 
Then, we us show that $\mathcal{Q}_t(x_{t-1},\xi_{[t]})$ is convex in $x_{t-1}$. From Equation \ref{eqn3},
\[
\mathcal{Q}_t(x_{t-1},\xi_{[t]})=\inf_{x_t \in \mathcal{X}_t(x_{t-1},\xi_t)} \{ f_t(x_t,\xi_t)+Q_{t+1}(x_t,\xi_{[t]}) \}=\inf_{x_t}\psi(x_{t-1},x_t,\xi_{[t]}),
\]
where
\[
\psi(x_{t-1},x_t,\xi_{[t]})=f_t(x_t,\xi_t)+Q_{t+1}(x_t,\xi_{[t]})+\mathbf{I}_{\mathcal{X}_t(x_{t-1}, \xi_t)}(x_t),
\]
\[
\mathbf{I}_C(\cdot)= \begin{cases}
0 & if \quad \cdot \in C\\
+\infty & otherwise
\end{cases}
\]
Here, $f_t(x_t,\xi_t)$ is convex in $x_t$ by definition, and $Q_{t+1}(x_t,\xi_{[t]})$ is convex by the induction hypothesis. 
Let us extend the feasible region $\mathcal{X}_t(x_{t-1},\xi_t)$, which is a set of $x_t$'s given $x_{t-1}$, into the set of $(x_{t-1},x_t)$ pair. 
Then, it is a convex set, because it is an intersection of hyperplanes and sublevel sets of convex functions. 
Thus, $\mathbf{I}_{\mathcal{X}_t(x_{t-1}, \xi_t)}(x_t)$ is convex in $(x_{t-1},x_t)$. 
Since the sum of convex functions is again convex, $\psi(x_t,x_{t-1},\xi_{[t]})$ is convex in $(x_{t-1},x_t)$. Therefore, $\mathcal{Q}_t(x_{t-1},\xi_{[t]})$, an infimal projection of $\psi(x_t,x_{t-1},\xi_{[t]})$ with respect to $x_t$, is convex in $x_{t-1}$.

For $t=T$, $V_{t+1}=0$. Hence, $\mathcal{Q}_{T+1}$ is convex in $x_T$. 
By the mathematical induction, $\mathcal{Q}_{t+1}(x_t,\xi_{[t+1]})$ is convex with respect to $x_t$, for $t=T-1,\ldots,1$.
Under the stagewise independence assumption \ref{itm1}, the conditional expectation operator becomes redundant, i.e. $Q_{t+1}(x_t,\xi_{[t]})$ does not depend on $\xi_{[t]}$.
Therefore, $Q_{t+1}(x_t,\xi_{[t]})$ is equivalent to $Q_{t+1}(x_t)$.
Thus, $Q_{t+1}(x_t)$ is also convex.
\clearpage

\section{Details on Stochastic Dual Dynamic Programming} \label{appendix:B}
In this section, we present the details on SDDP.
SDDP algorithm obtains the optimal solution thought the following steps.

\textbf{1. Create scenario tree}

A scenario tree is generated by sampling random vectors $\xi_t^s$ for $j=1,\ldots,N_t$ where $N_t$ indicates the number of scenario nodes at state $t$ from the original distribution $P_t$ for $t=2,\ldots,T$.

\textbf{2. Forward step}

From $t=1$ to $T$, obtain current optimal solutions $x_t^*$ of the $n$-th iteration of stage $t$ by solving the following stagewise subproblems sequentially given random vectors, $\xi_t^s$.
\begin{align} \label{eqn5}
x_t^*=\underset{x_t \in \mathcal{X}_{t}(x_{t-1}, \xi_t^s)}{\arg\min} f_t(x_t, \xi_t^s)+Q_{t+1}^n(x_t).
\end{align}

The value function approximated as a piecewise linear convex function in SDDP. 
Thus, the current value function $Q_{t+1}^n$ is expressed by $\max_{k}\{(\beta_{t+1}^k)^\top x_t+\alpha_{t+1}^k\}$, a function of subgradient cutting planes (cuts) constructed in the backward steps.
Then, we can reformulate the subproblem (\ref{eqn5}) without value function as follows.
\begin{align} \label{eqn6}
    \underset{x_t \in \mathcal{X}_{t}(x_{t-1}, \xi_t^s)}{\min} \quad &f_t(x_t, \xi_t^s)+\theta_{t+1}, \notag \\ 
    s.t \quad &\theta_{t+1} \geq (\beta_{t+1}^k)^\top x_t+\alpha_{t+1}^k, \forall k=1,\ldots,n.
\end{align}

By performing the forward step for the $M$ sampled set of scenarios, calculate estimator of optimal objective value as follows:

\[
    \Bar{v} = \frac{1}{M} \sum_{s=1}^M \sum_{t=1}^T f_t(x_t^*, \xi_t^s), \quad \sigma_{\Bar{v}}^2 = \frac{1}{M-1} \sum_{s=1}^M \bigg[\sum_{t=1}^T f_t(x_t^*, \xi_t^s) - \Bar{v}\bigg]
\]

This gives $100(1-\alpha)$\% confidence interval for the optimal objective value, and we use $\Bar{v} + z_{\frac{\alpha}{2}} \sigma_{\Bar{v}}^2 / \sqrt{M}$ as an upper bound for the optimal value. 

\textbf{3. Backward step}

In backward step from $t=T$, the approximated value function $Q_t^n$ is updated by adding linear cut derived from optimal solutions of stage $t$ subproblem.
For the $n$-th iteration of stage $t$ with sample $\xi_t^s$, the approximated value function $Q_t^n$ is updated to $Q_t^{n+1}$ using primal and dual optimal solutions $(x_t^*, u_t^*, v_t^*)$ of the stage $t$ subproblem (\ref{eqn6}) as follows.
\[
Q_t^{n+1}(x_{t-1})=\max\{Q_t^n(x_{t-1}),(\beta_t^{n+1})^\top x_{t-1} + \alpha^{n+1}_t\}
\]
where
\begin{align*}
    \beta_t^{n+1} & =\frac{1}{N_t}\sum_{s=1}^{N_t}\bigg[\sum_{i=1}^{p_t}u^\ast_{t,i}\nabla_{x_{t-1}}h_{t,i}(x_{t-1},\xi_t^s) +\sum_{j=1}^{q_t}v^\ast_{t,j}\nabla_{x_{t-1}}b_{t,j}(x_{t-1},\xi_t^s)\bigg].
\end{align*}
\[
\alpha_t^{n+1}=\frac{1}{N_t}\sum_{s=1}^{N_t} f_t(x^\ast_t,\xi_t^s)+V^n_{t+1}(x^\ast_t).
\]
We can then compute a lower bound from the updated value function.
That is, $\min_{x \in \mathcal{X}} [f_1(x_1)+Q_2^{n+1}(x_1)]$ is used as a lower bound for the optimal value.

\textbf{4. Stopping criterion}

We can use the difference between the upper bound and the lower bound as the stopping criterion for SDDP algorithm.
Once this difference is sufficiently small, we can conclude that the algorithm has converged to the optimal value.
The forward and backward step are repeated until the stopping criterion is satisfied.

In summary, 1) Terminate the SDDP algorithm if $|upper bound - lower bound| \leq threshold$, 2) Otherwise, return to the step 2, and proceed with the forward and backward step again.
\clearpage

\section{Architecture of TranSDDP and TranSDDP-Decoder} \label{appendix:C}
In this section, we present the architecture of TranSDDP and its variant, TranSDDP-Decoder, which omits the encoder part.

\begin{figure*}[h]
    \centering
    \includegraphics[width=\textwidth]{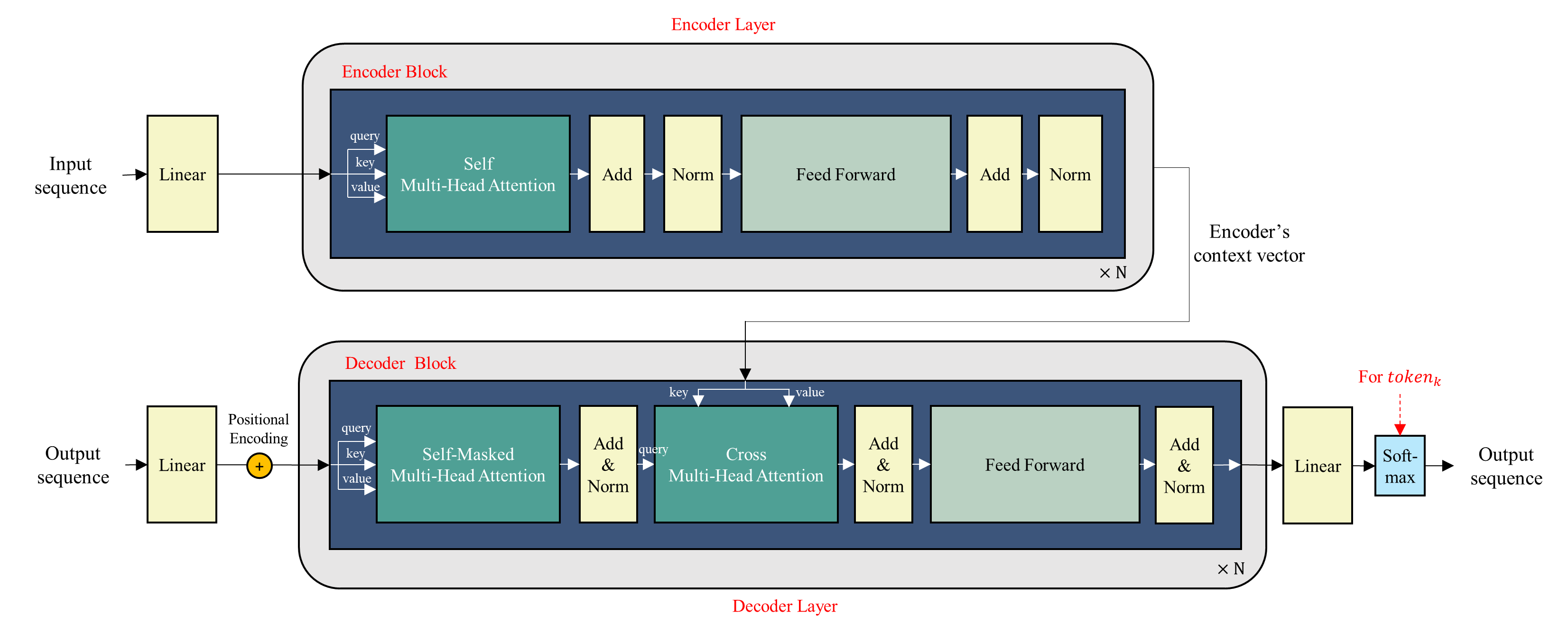}
    \caption{Architecture of TranSDDP}
    \label{fig3}
\end{figure*}

\begin{figure*}[h]
    \centering
    \includegraphics[width=\textwidth]{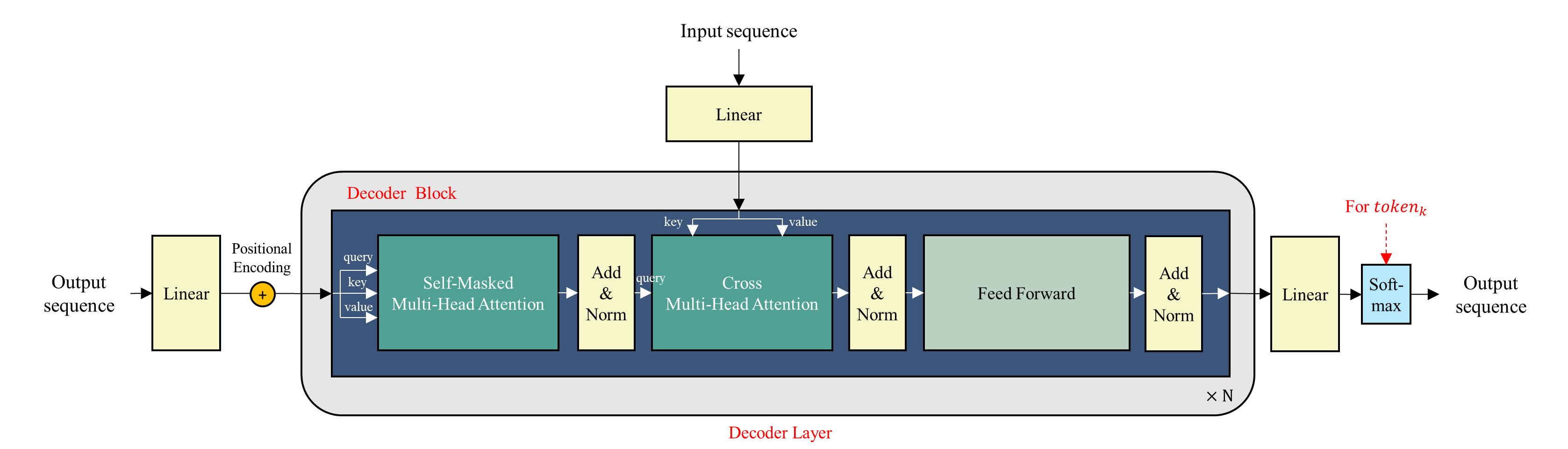}
    \caption{Architecture of TranSDDP-Decoder}
    \label{fig4}
\end{figure*}
\clearpage

\section{Design of the learning system} \label{appendix:D}
In this section, we detail the procedures for training, validation, and test using the TranSDDP and TranSDDP-Decoder algorithms.
Validation process is similar to the training process outlined in Section \ref{section3.3}.
The performance of the models is evaluated using the results of the inference step during test.
\subsection{Training system}
\begin{figure*}[h]
    \centering
    \includegraphics[width=\textwidth]{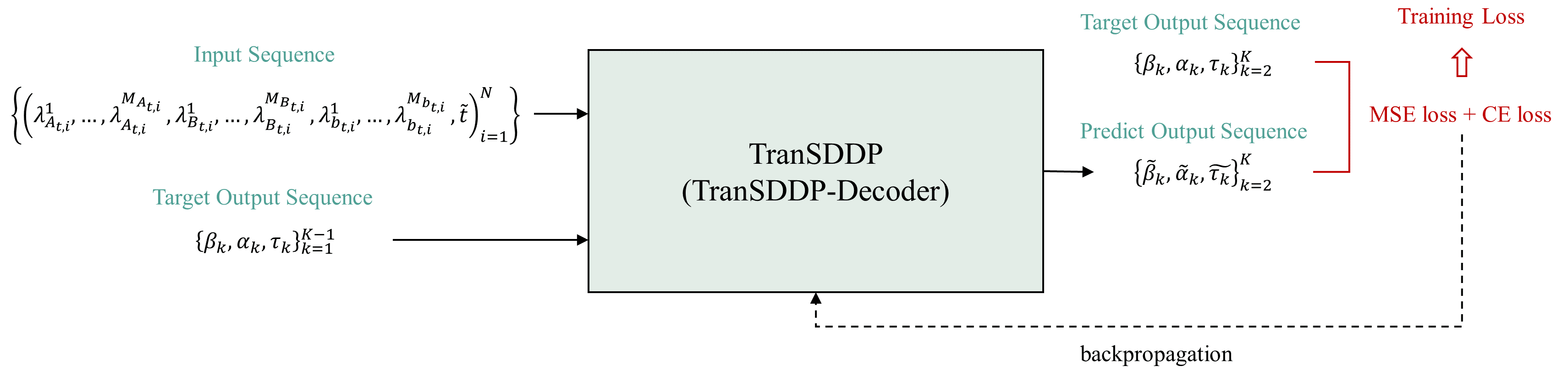}
    \vspace{-0.3cm}
    \caption{Design of training system}
    \label{fig5}
\end{figure*}

\subsection{Validation system}
\begin{figure*}[h]
    \centering
    \includegraphics[width=\textwidth]{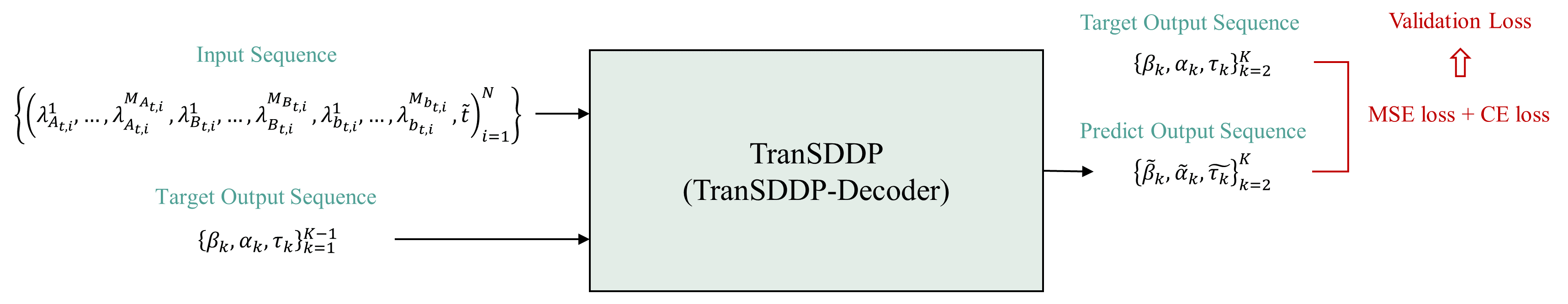}
    \vspace{-0.3cm}
    \caption{Design of validation system}
    \label{fig6}
\end{figure*}

\subsection{Test system}
In the testing phase, the TranSDDP algorithm (or its variance, TranSDDP-Decoder) generates cuts iteratively by taking the input sequence and previously generated cuts.
Specifically, it generates the $k$-th cut, $\{\Tilde{\beta_k}, \Tilde{\alpha_k}\}$, by utilizing the input sequence and previous cuts, $\{\Tilde{\beta_i}, \Tilde{\alpha_i}\}_{i=1}^{k-1}$.
The process continues until the end of the sequence is signaled by the end token, i.e., $\tau_k=(0, 0, 0, 1)$.
The performance of TranSDDP and TranSDDP-Decoder is evaluated by comparing the objective values obtained from solving the multistage stochastic optimization problem formulated using the cuts generated by the algorithms.
\begin{figure*}[h]
    \centering
    \includegraphics[width=\textwidth]{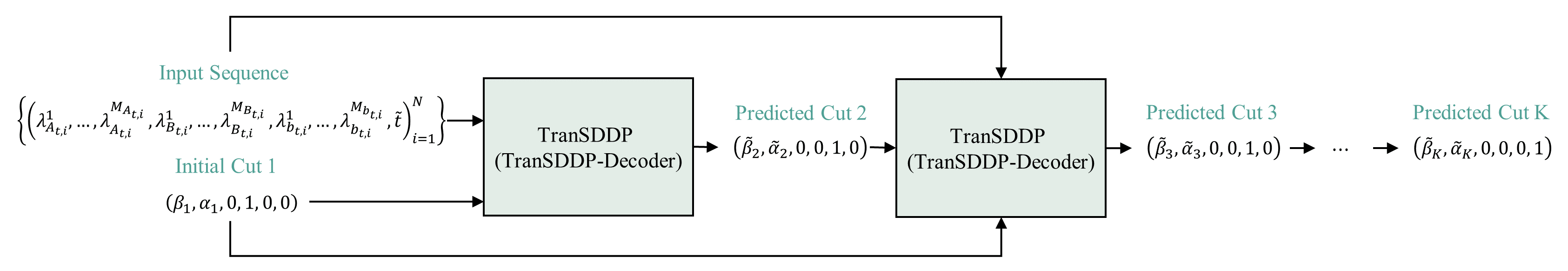}
    \vspace{-0.3cm}
    \caption{Design of test system}
    \label{fig7}
\end{figure*}
\clearpage

\section{Formulation of numerical experiments}
In this section, we provide a details of the decision variables, parameters, and stagewise subproblem formulations for the energy planning, financial planning, and production planning problems presented in Section \ref{section4}.

\subsection{Energy Planning} \label{appendix:E.1}

\begin{itemize}
    \item Stage $1$ subproblem
    \begin{alignat*}{5}
        &\text{minimize} &\quad& c_1^W W_1+c_1^H H_1+e^{-a_1 r_1^{final}+b_1}+Q_2(r_1^{final}) && \\
        &\text{subject to} &\quad& r_1^{init}=r_0^{init} &&\quad \text{Initial reservoir} \\
        & &\quad& r_1^{final}=r_1^{init}-W_1 &&\quad \text{Reservoir balance} \\
        & &\quad& W_1+H_1 \geq d_1 &&\quad \text{Demand} \\ 
        & &\quad& r_1^{final},W_1,H_1 \geq 0 &&\quad \text{Non-negativity}
    \end{alignat*}
    
    \item Stage $t$ subproblem ($t=2,\ldots,T-1$)
    \begin{alignat*}{5}
        &\text{minimize} &\quad& c_t^W W_t+c_t^H H_t+e^{-a_t r_t^{final}+b_t}+Q_{t+1}(r_t^{final}) && \\
        &\text{subject to} &\quad& r_t^{init}=r_{t-1}^{final}+I_t &&\quad \text{Initial reservoir} \\
        & &\quad& r_t^{final}=r_t^{init}-W_t &&\quad \text{Reservoir balance} \\
        & &\quad& W_t+H_t \geq d_t &&\quad \text{Demand} \\ 
        & &\quad& r_t^{final},W_t,H_t \geq 0 &&\quad \text{Non-negativity}
    \end{alignat*}
    
    \item Stage $T$ subproblem
    \begin{alignat*}{5}
        &\text{minimize} &\quad& c_T^W W_T+c_T^H H_T+e^{-a_T r_T^{final}+b_T} && \\
        &\text{subject to} &\quad& r_T^{init}=r_{T-1}^{final}+I_T &&\quad \text{Initial reservoir} \\
        & &\quad& r_T^{final}=r_T^{init}-W_T &&\quad \text{Reservoir balance} \\
        & &\quad& W_T+H_T \geq d_T &&\quad \text{Demand} \\ 
        & &\quad& r_T^{final},W_T,H_T \geq 0 &&\quad \text{Non-negativity}
    \end{alignat*}
\end{itemize}

\begin{table}[hb]
\renewcommand{\arraystretch}{1.05}
\caption{Decision variables and parameters for energy planning problem}
\label{tab5}
\vskip 0.1in
\centering
\resizebox{0.75\textwidth}{!}{
    \begin{tabular}{lll}
    \toprule
    \textbf{Decision variables} & \multicolumn{2}{l}{\textbf{Description}} \\
    \midrule
    $r_t^{init}$ & Water reservoir level in the beginning of stage $t$ \\
    $r_t^{final}$ & Water reservoir level in the end of stage $t$ \\
    $W_t$ & Hydro electricity generation level at stage $t$ \\
    $H_t$ & Thermal electricity generation level at stage $t$ \\
    \toprule
    \textbf{Parameters} & \textbf{Description} & \textbf{Value} \\
    \midrule
    $r_0^{init}$ & Initial water reservoir level & 40 \\
    $c_t^W$ & Cost of hydro electricity production per unit at stage $t$ & 2 \\
    $c_t^H$ & Cost of thermal electricity production per unit at stage $t$ & 7 \\
    $d_t$ & Electricity demand at stage $t$ & 20 \\
    $a_t$ & Reservoir level utility coefficient & 0.1 \\
    $b_t$ & Reservoir level utility scaling constant & 5 \\
    $I_t$ & Water inflow to reservoir in the beginning of stage $t$ & $I_t \sim \mathcal{N}(\mu_I, \sigma_I^2)$ \\ 
    $\mu_I$ & Mean of water inflow & $\mu_I \sim \mathcal{U}(15, 25)$ \\ 
    $\sigma_I$ & Standard deviation of water inflow & $\sigma_I \sim \mathcal{U}(4, 6)$ \\
    \bottomrule
    \end{tabular}
    }
\end{table}
\clearpage

\subsection{Financial planning} \label{appendix:E.2}

\begin{itemize}
    \item Stage $1$ subproblem
    \begin{alignat*}{5}
        &\text{minimize} &\quad& -U(C_1) + Q_2(S_1, B_1) && \\
        &\text{subject to} &\quad& S_1 + B_1 + C_1 = w^{init} &&\quad \text{Initial wealth} \\
        & &\quad& S_1, B_1, C_1 \geq 0 &&\quad \text{Non-negativity}
    \end{alignat*}
    
    \item Stage $t$ subproblem ($t=2,\ldots,T-1$)
    \begin{alignat*}{5}
        &\text{minimize} &\quad& -U(C_t) + Q_{t+1}(S_t, B_t) && \\
        &\text{subject to} &\quad& S_t + B_t + C_t = r^{free} \Delta t B_{t-1} + r^{stock}S_{t-1} &&\quad \text{Wealth balance} \\
        & &\quad& S_t, B_t, C_t \geq 0 &&\quad \text{Non-negativity}
    \end{alignat*}
    
    \item Stage $T$ subproblem
    \begin{alignat*}{5}
        &\text{minimize} &\quad& -U(C_T) - U(W_T) && \\
        &\text{subject to} &\quad& W_T + S_T = r^{free} \Delta t B_{T-1} + r^{stock}S_{T-1} &&\quad \text{Wealth balance} \\
        & &\quad& S_T, B_T, C_T, W_T \geq 0 &&\quad \text{Non-negativity}
    \end{alignat*}

    \item[] where U is the isoelastic utility function defined as follows:
    \[
    U(x) = \begin{cases}
    log \, x & if \, \eta=1, \\
    \frac{1}{1-\eta}x^{1-\eta} & if \, 0 \leq \eta < 1.
    \end{cases}
    \]
\end{itemize}

\begin{table}[hb]
\renewcommand{\arraystretch}{1.05}
\caption{Decision variables and parameters for financial planning problem}
\label{tab6}
\vskip 0.1in
\centering
\resizebox{0.75\textwidth}{!}{
    \begin{tabular}{lll}
    \toprule
    \textbf{Decision variables} & \multicolumn{2}{l}{\textbf{Description}} \\
    \midrule
    $S_t$ & Amount invested into stock at stage $t$ \\
    $B_t$ & Amount invested into bond at stage $t$ \\
    $C_t$ & Consumption at stage $t$ \\
    $W_t$ & Wealth in the beginning of stage $t$ \\
    \toprule
    \textbf{Parameters} & \textbf{Description} & \textbf{Value} \\
    \midrule
    $w^{init}$ & Initial wealth & 100 \\
    $\eta$ & Utility risk aversion coefficient & 1 \\
    $r^{free}$ & Risk-free rate of bond & 1.03 \\
    $r^{stock}$ & Rate of return of stock & $\log(r^{stock}) \sim \mathcal{N}((\mu-{\sigma^2}/2)\Delta t, \sigma^2 \Delta t)$ \\ 
    $\mu$ & Expected return of stock & $\mu \sim \mathcal{U}(0.04, 0.08)$ \\
    $\sigma$ & Volatility of stock & $\sigma \sim \mathcal{U}(0.15, 0.25)$ \\
    \bottomrule
    \end{tabular}
    }
\end{table}
\clearpage

\subsection{Production Planning} \label{appendix:E.3}

\begin{itemize}
    \item Stage $1$ subproblem
    \begin{alignat*}{5}
        &\text{minimize} &\quad& \sum_{i\in I} y_{1,i}b_{1,i}+\sum_{i\in I} s_{1,i}c_{1,i}+Q_2(s_1) && \\
        &\text{subject to} &\quad& \sum_{i\in I} x_{1,i}a_{1,i} \leq r_1 &\quad \forall i\in I &\quad \text{Resource limit} \\
        & &\quad& s_{1,i}=x_{1,i}+y_{1,i} &\quad \forall i\in I &\quad \text{Storage balance} \\
        & &\quad& x_{1,i},y_{1,i},s_{1,i} \geq 0 &\quad \forall i\in I &\quad \text{Non-negativity} \\ 
    \end{alignat*}
    
    \item Stage $t$ subproblem ($t=2,\ldots,T-1$)
    \begin{alignat*}{5}
        &\text{minimize} &\quad& \sum_{i\in I} y_{t,i}b_{t,i}+\sum_{i\in I} s_{t,i}c_{t,i}+Q_{t+1}(s_t) && \\
        &\text{subject to} &\quad& \sum_{i\in I} x_{t,i}a_{t,i} \leq r_t &\quad \forall i\in I &\quad \text{Resource limit} \\
        & &\quad& s_{t,i}=s_{t-1,i}+x_{t,i}+y_{t,i}-d_{t,i} &\quad \forall i\in I &\quad \text{Storage balance} \\
        & &\quad& x_{t,i},y_{t,i},s_{t,i} \geq 0 &\quad \forall i\in I &\quad \text{Non-negativity} \\ 
    \end{alignat*}
    
    \item Stage $T$ subproblem
    \begin{alignat*}{5}
        &\text{minimize} &\quad& \sum_{i\in I} y_{T,i}b_{T,i} && \\
        &\text{subject to} &\quad& \sum_{i\in I} x_{T,i}a_{T,i} \leq r_T &\quad \forall i\in I &\quad \text{Resource limit} \\
        & &\quad& s_{T,i}=s_{T-1,i}+x_{T,i}+y_{T,i}-d_{T,i} &\quad \forall i\in I &\quad \text{Storage balance} \\
        & &\quad& x_{T,i},y_{T,i},s_{T,i} \geq 0 &\quad \forall i\in I &\quad \text{Non-negativity} \\ 
    \end{alignat*}
\end{itemize}

\vspace{-0.5cm}
\begin{table}[hb]
\renewcommand{\arraystretch}{1.03}
\caption{Decision variables and parameters for production planning problem}
\label{tab7}
\vskip 0.1in
\centering
\resizebox{0.75\textwidth}{!}{
    \begin{tabular}{lll}
    \toprule
    \textbf{Decision variables} & \multicolumn{2}{l}{\textbf{Description}} \\
    \midrule
    $x_{t, i}$ & Quantity of product $i$ produced at stage $t$ \\
    $y_{t, i}$ & Quantity of product $i$ outsourced at stage $t$ \\
    $s_{t, i}$ & Quantity of product $i$ stored at stage $t$ \\
    \toprule
    \textbf{Parameters} & \textbf{Description} & \textbf{Value} \\
    \midrule
    $i$ & Product number & $(1, 2, 3)$ \\
    $a_{t, i}$ & Production cost of product $i$ at stage $t$ & $(1, 2, 5)$ \\
    $b_{t, i}$ & Outsourcing cost of product $i$ at stage $t$ & $(6, 12, 20)$ \\
    \multirow{2}{*}{$c_{t, i}$} & Storage cost of product $i$ from the end of stage $t$ & \multirow{2}{*}{$(3, 7, 10)$} \\
    & to beginning of stage $t+1$ & \\
    $r_t$ & Maximum production resource available at stage $t$ & 10 \\
    $d_{t, i}$ & Random demand of product $i$ at stage $t$ & $d_{i, t} \sim \mathcal{N}(\mu_{d_i}, {\sigma_{d_i}}^2)$ where\\
    $\mu_{d_1}$ & Mean demand of product 1 & $\mu_{d_1} \sim \mathcal{U}(3, 6)$ \\
    $\sigma_{d_1}$ & Standard deviation of demand of product 1 & $\sigma_{d_1} \sim \mathcal{U}(0.2, 0.4)$ \\
    $\mu_{d_2}$ & Mean demand of product 2 & $\mu_{d_2} \sim \mathcal{U}(1.5, 4)$ \\
    $\sigma_{d_2}$ & Standard deviation of demand of product 2 & $\sigma_{d_2} \sim \mathcal{U}(0.1, 0.2)$ \\
    $\mu_{d_3}$ & Mean demand of product 3 & $\mu_{d_3} \sim \mathcal{U}(1, 2)$ \\
    $\sigma_{d_3}$ & Standard deviation of demand of product 3 & $\sigma_{d_3} \sim \mathcal{U}(0.05, 0.1)$ \\
    \bottomrule
    \end{tabular}
    }
\end{table}
\clearpage

\section{Additional experimental results}\label{appendix:F}
In this section, we present additional experimental results for solving 7-stage problems using TranSDDP and TranSDDP-Decoder.
These results include the comparison of the value function and its approximations, as well as results obtained from the training and validation process.

In multistage stochastic programming problems, accurately representing the true value function is a significant challenging.
This is because the value function at earlier stages relies on approximations derived from the value function at later stages.
However, since there is no value function in the final stage, the objective function at stage $T$ can be regarded as the true value function (referred to as $Q_T$).
This enables the comparison between the value function and its approximations.

In the graph illustrating the comparison between the value function and its approximation, it is evident that TranSDDP and TranSDDP-Decoder, which utilize the information of distribution in which stochastic elements are sampled, provide a single value function approximation for the corresponding distribution. 
Conversely, for SDDP, which approximates the value function through random sampling, there is noticeable variability observed in the results.

\subsection{Energy planning} \label{appendix:F.1}
\subsubsection{Comparison of the value function and its approximations for energy planning problem} \label{appendix:F.1.1}
This graph presents the results of an experiment conducted by sampling the stochastic element, water inflow, from $\mathcal{N}(20, 5)$, a total of 100 times.
The shaded area represents the variance arising from the sampling, and the solid line represents its mean value.

\begin{figure}[h]
    \centering
    \includegraphics[width=\textwidth]{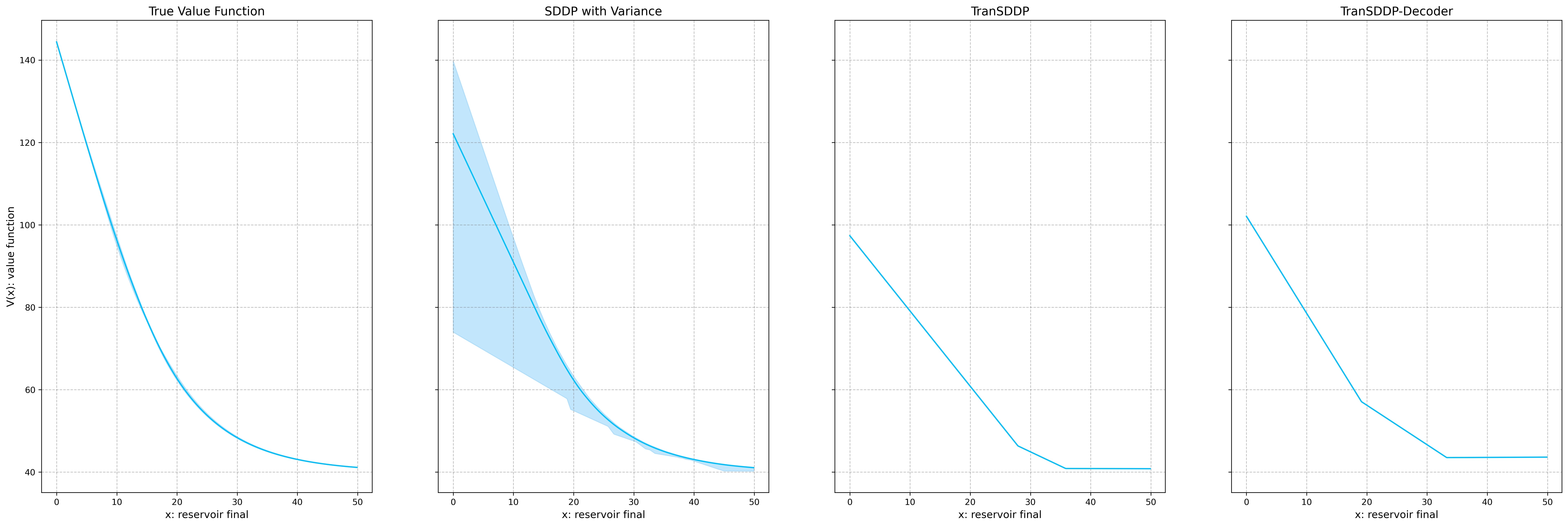}
    \vspace{-0.3cm}
    \caption{Comparison of the value function and its approximations for energy planning problem}
\end{figure}
\clearpage

\subsubsection{Results of training and validation process for energy planning problem} \label{appendix:F.1.2}

\begin{figure}[h]
    \centering
    \subfigure[Error ratio]{
    \includegraphics[width=0.4\textwidth]{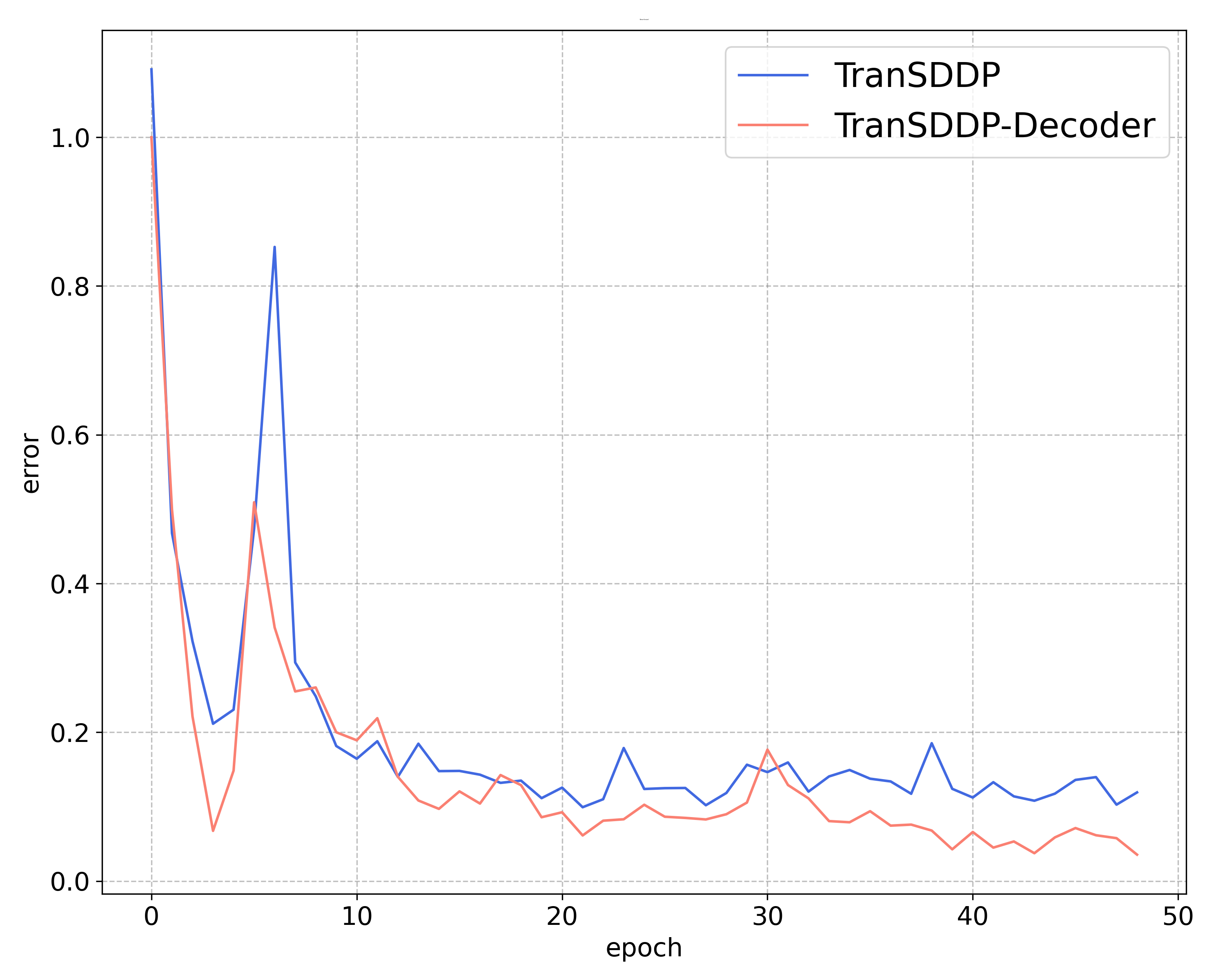}
    }
    \subfigure[Loss]{
    \includegraphics[width=0.4\textwidth]{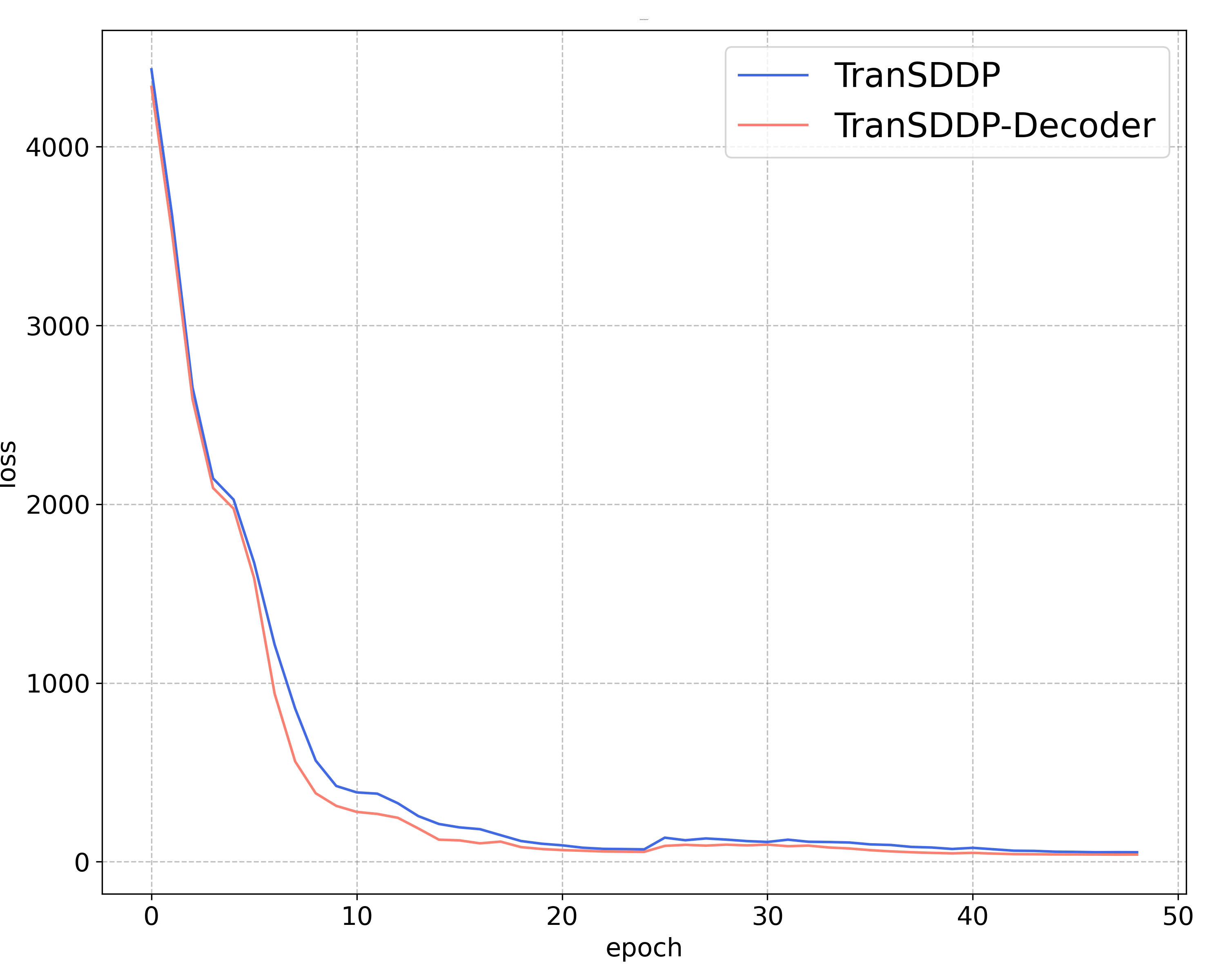}
    }
    \caption{
    Training results for energy planning problem
    }
\end{figure}

\begin{figure}[h]
    \centering
    \subfigure[Error ratio]{
    \includegraphics[width=0.4\textwidth]{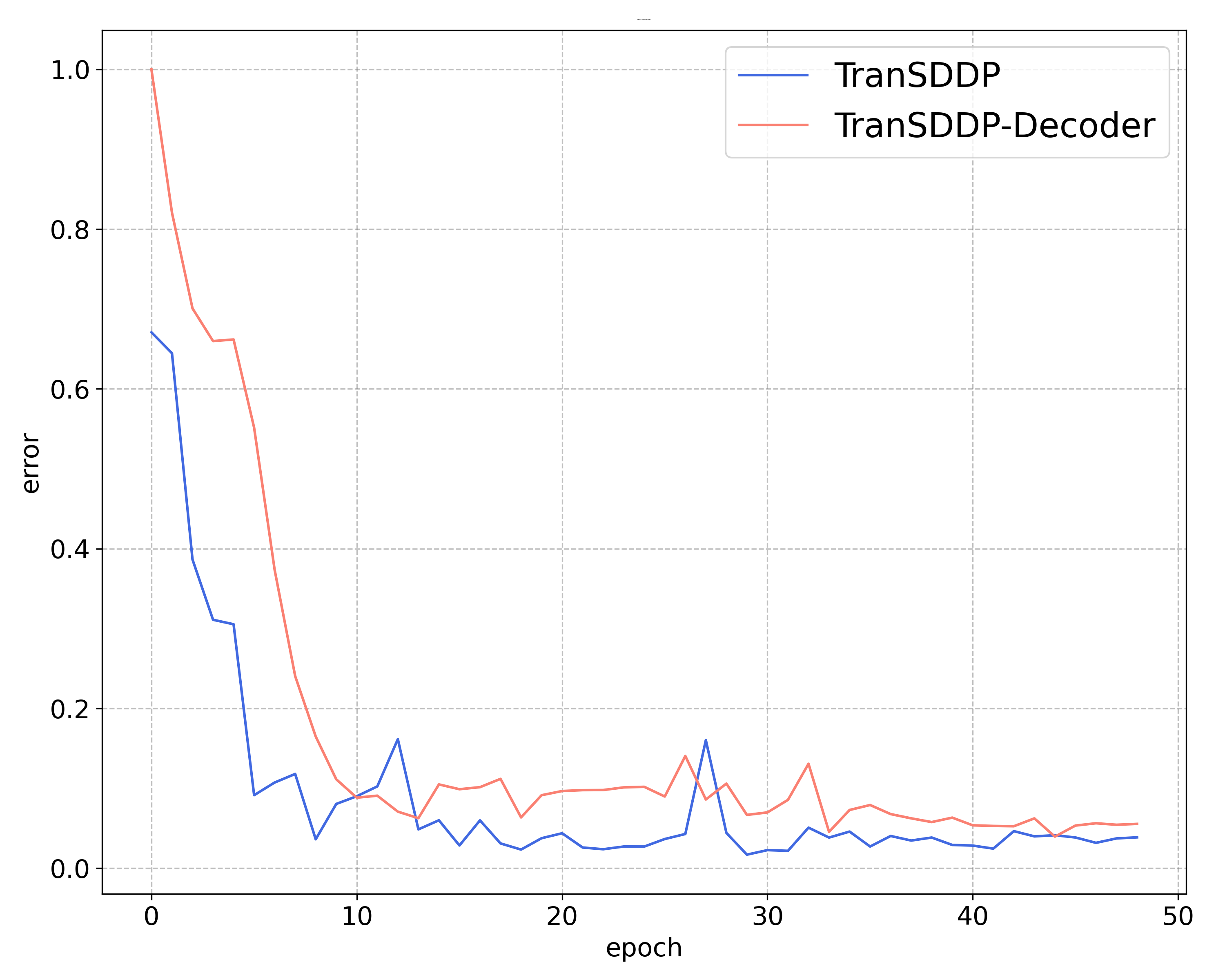}
    }
    \subfigure[Loss]{
    \includegraphics[width=0.4\textwidth]{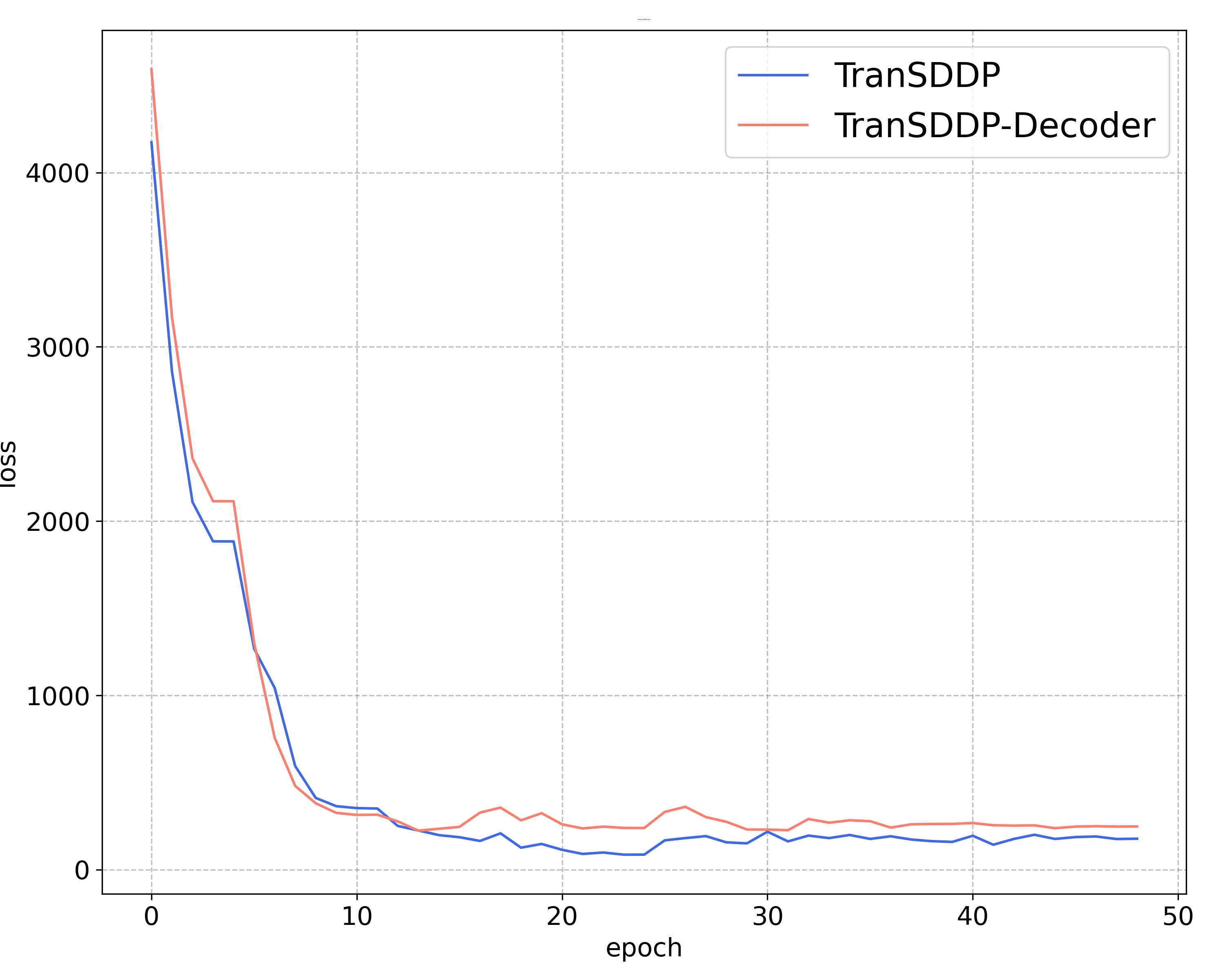}
    }
    \caption{
    Validation results for energy planning problem
    }
\end{figure}
\clearpage

\subsection{Financial planning} \label{appendix:F.2}
\subsubsection{Comparison of the value function and its approximations for financial planning problem} \label{appendix:F.2.1}
This graph presents the results of an experiment conducted by sampling the stochastic element, rate of return of stock $\log r^{stock}$, from $\mathcal{N}((\mu-{\sigma^2}/2)\Delta t, \sigma^2 \Delta t)$ with parameter $\mu=0.06$ and $\sigma=0.2$, a total of 100 times.
The upper row graph represents the value function for variable $S_t$, whereas the lower row graph represents the value function for variable $B_t$.

\begin{figure*}[h]
    \centering
    \includegraphics[width=\textwidth]{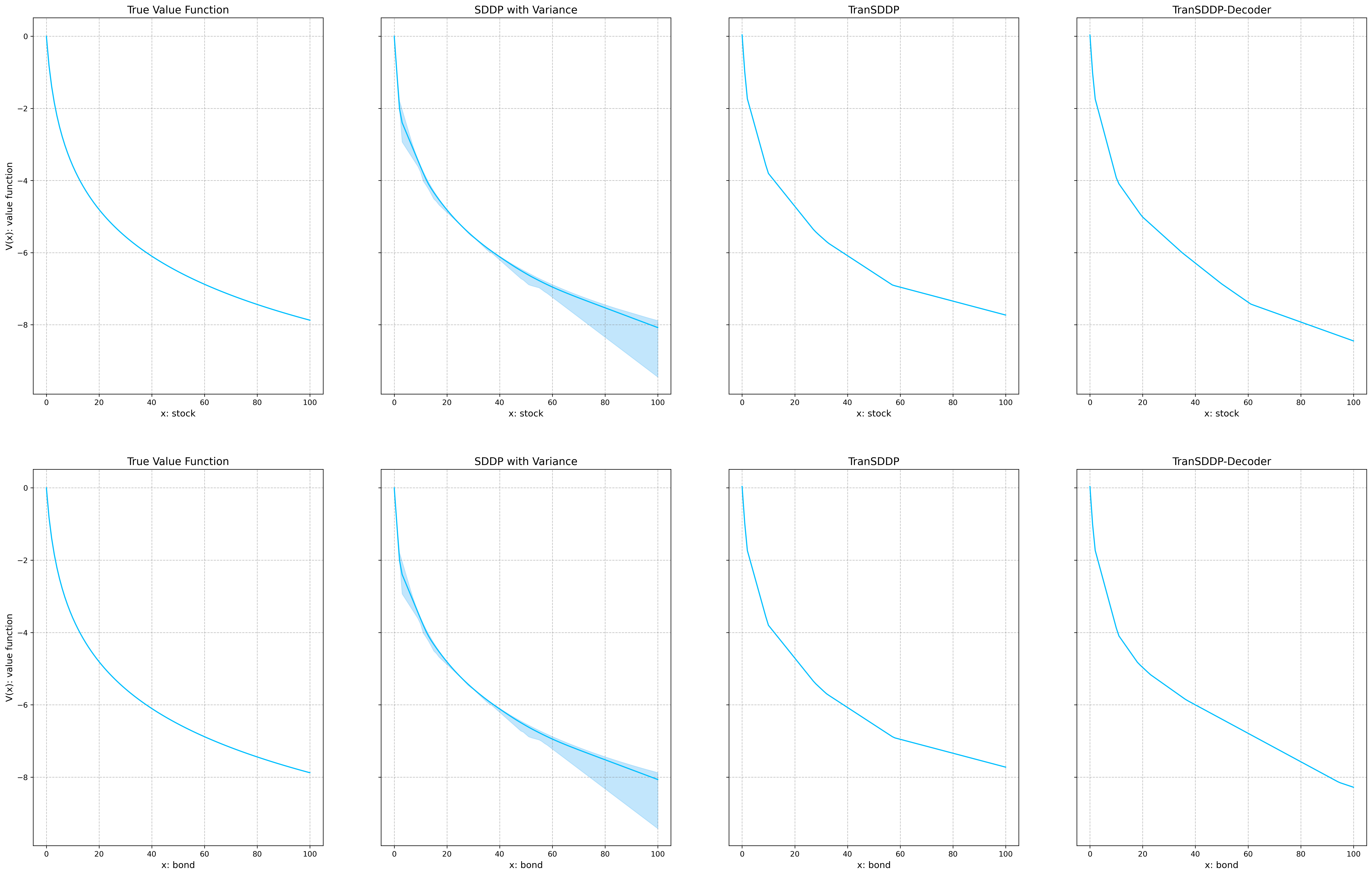}
    \vspace{-0.3cm}
    \caption{Comparison of the value function and its approximations for financial planning problem}
\end{figure*}
\clearpage

\subsubsection{Results of training and validation process for financial planning problem} \label{appendix:F.2.2}

\begin{figure}[h]
    \centering
    \subfigure[Error ratio]{
    \includegraphics[width=0.4\textwidth]{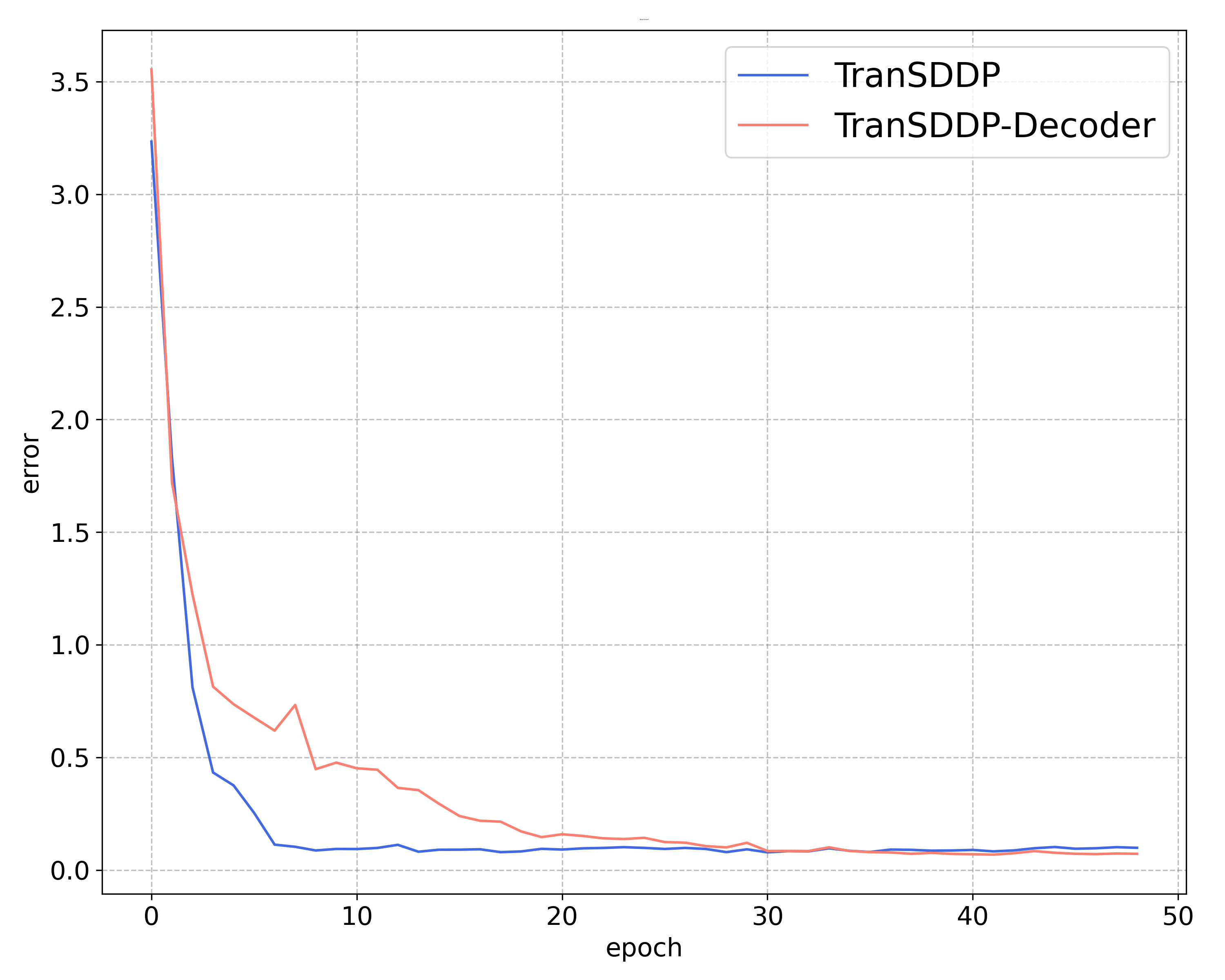}
    }
    \subfigure[Loss]{
    \includegraphics[width=0.4\textwidth]{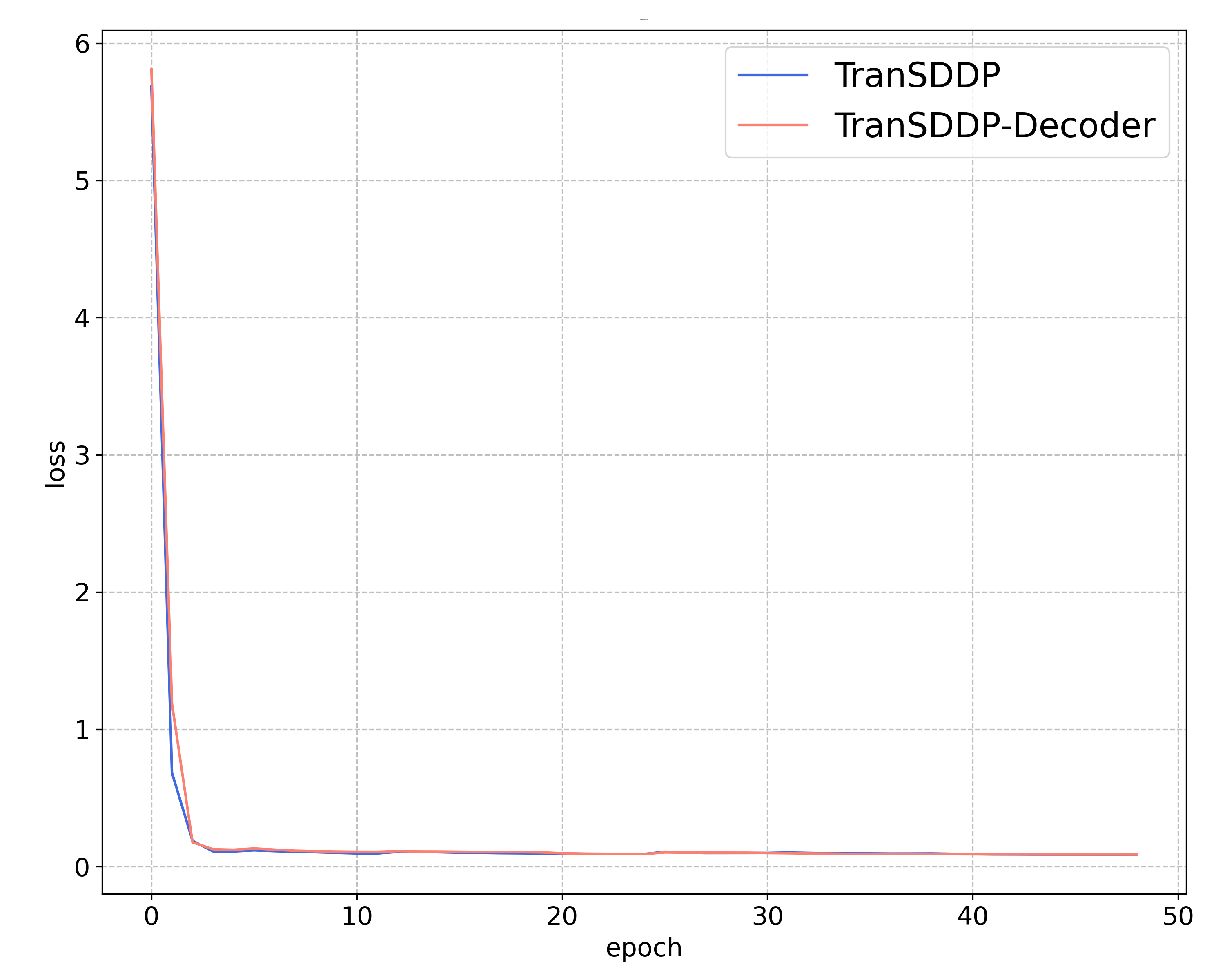}
    }
    \caption{
    Training results for financial planning problem
    }
\end{figure}

\begin{figure}[h]
    \centering
    \subfigure[Error ratio]{
    \includegraphics[width=0.4\textwidth]{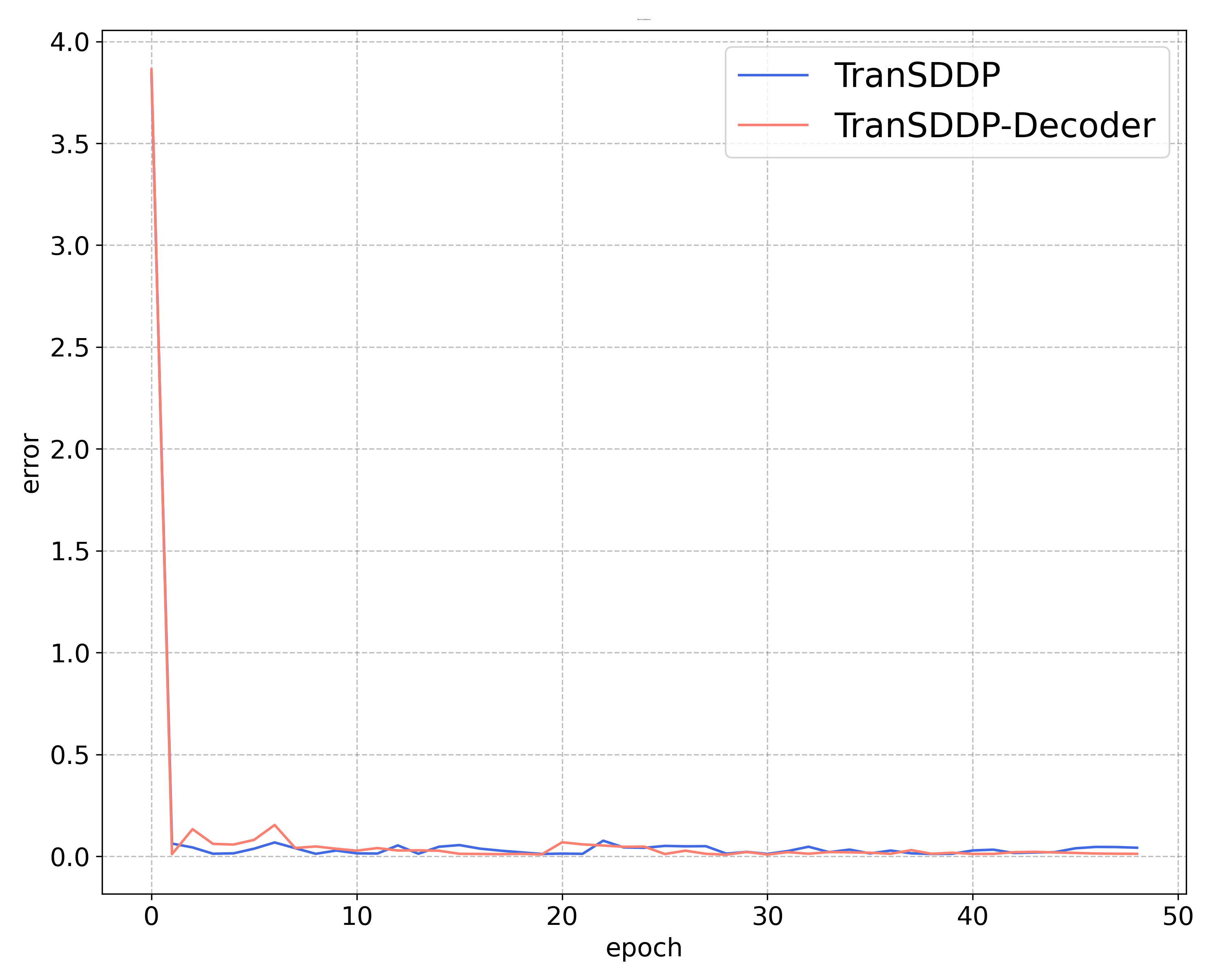}
    }
    \subfigure[Loss]{
    \includegraphics[width=0.4\textwidth]{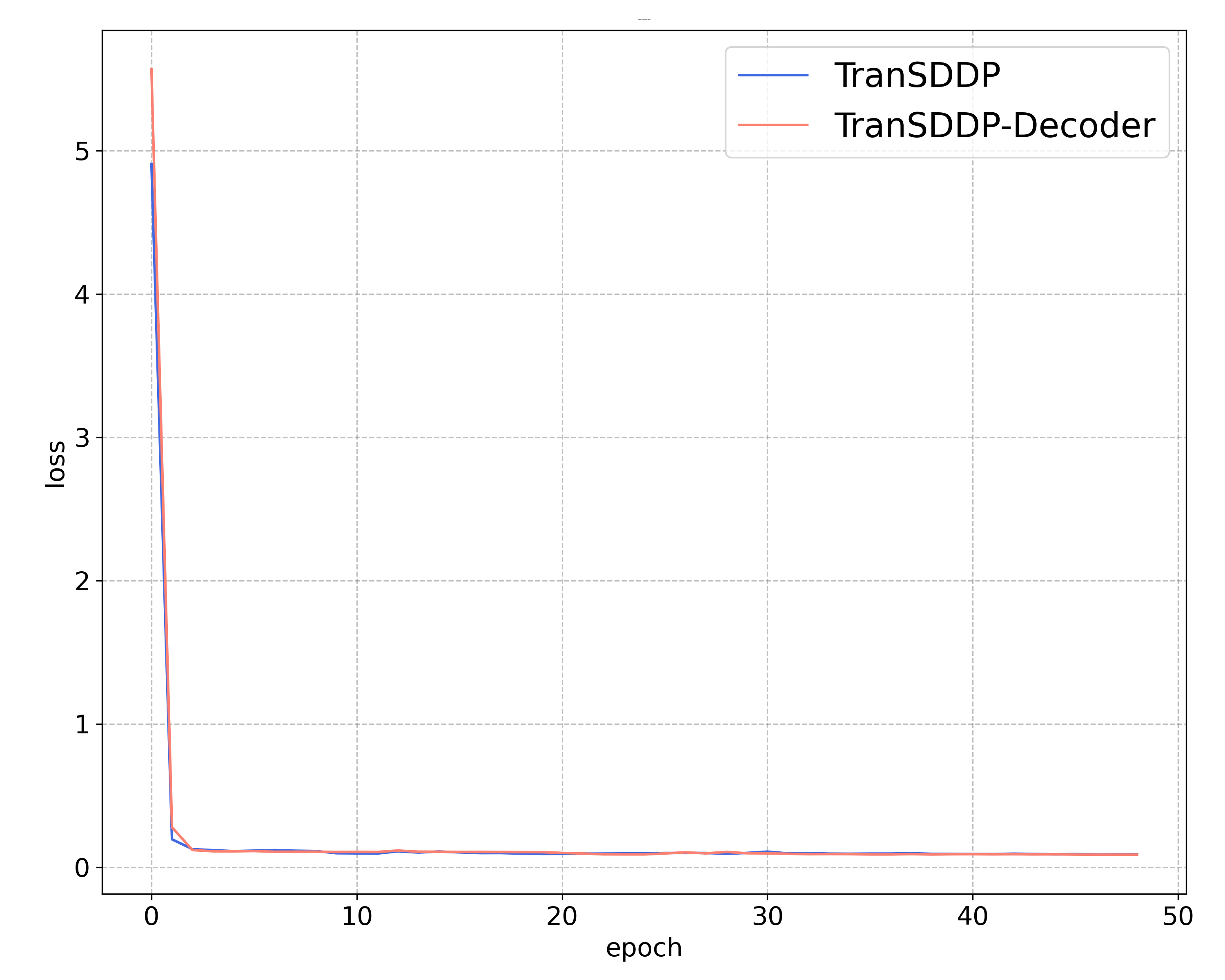}
    }
    \caption{
    Validation results for financial planning problem
    }
\end{figure}
\clearpage

\subsection{Production planning} \label{appendix:F.3}
\subsubsection{Comparison of the value function and its approximations for production planning problem} \label{appendix:F.3.1}
This graph presents the results of an experiment conducted by sampling the stochastic element, demand of product $i$ at stage $t$, $d_{t, i}$, that follows $\mathcal{N}(\mu_{d_i}, {\sigma_{d_i}}^2)$ with parameter $\mu_{d_1}=4.5$, $\sigma_{d_1}=0.3$, $\mu_{d_2}=2.75$, $\sigma_{d_2}=0.15$, $\mu_{d_3}=1.5$, $\sigma_{d_3}=0.075$, a total of 100 times.
The upper row graph represents the value function for variable $s_{t, 1}$, the middle row graph represents the value function for variable $s_{t, 2}$, and the lower row graph represents the value function for variable $s_{t, 3}$.

\begin{figure*}[h]
    \centering
    \includegraphics[width=\textwidth]{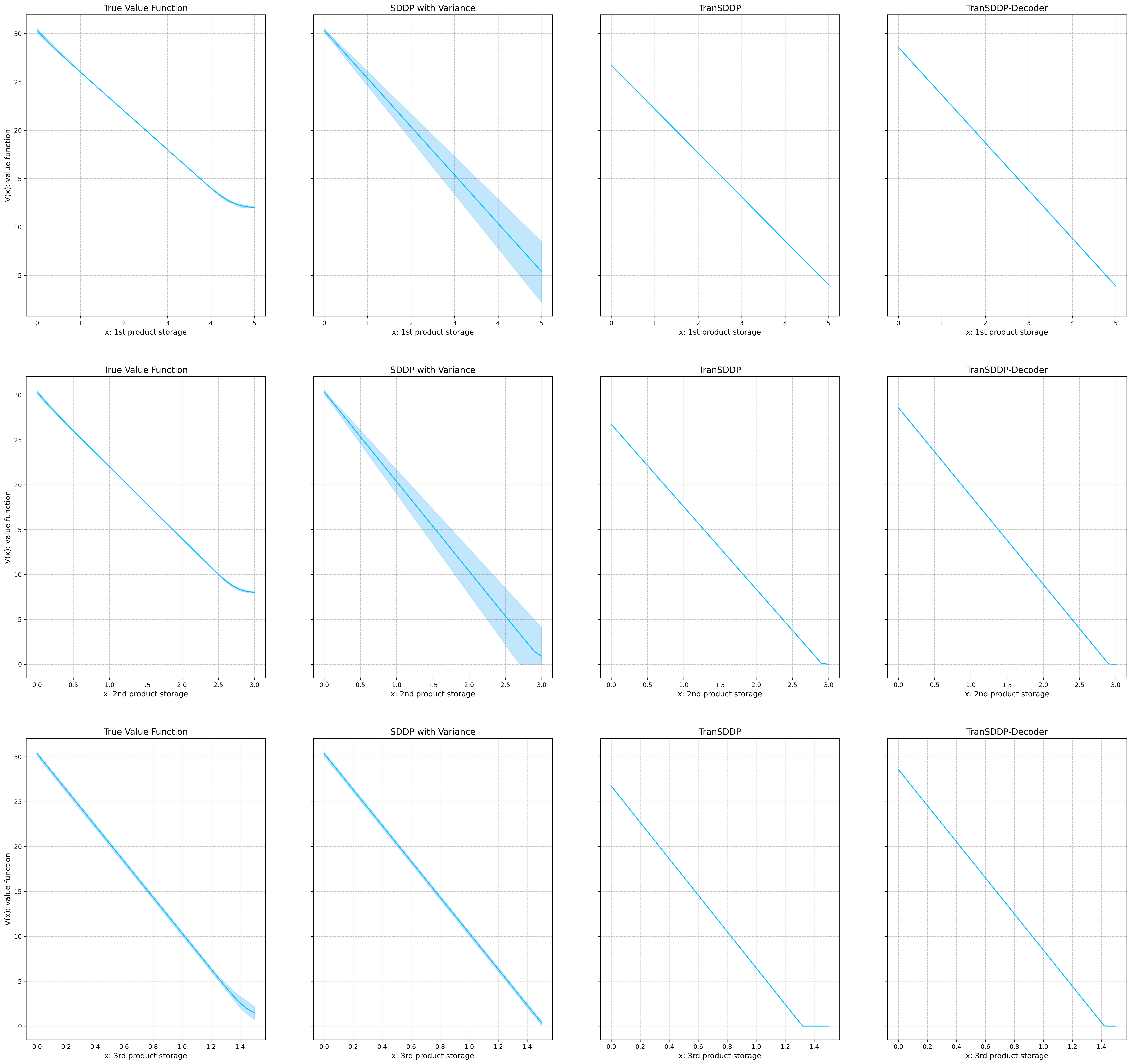}
    \vspace{-0.3cm}
    \caption{Comparison of the value function and its approximations for production planning problem}
\end{figure*}
\clearpage

\subsubsection{Results of training and validation process for production planning problem} 
\begin{figure}[h]
    \centering
    \subfigure[Error ratio]{
    \includegraphics[width=0.4\textwidth]{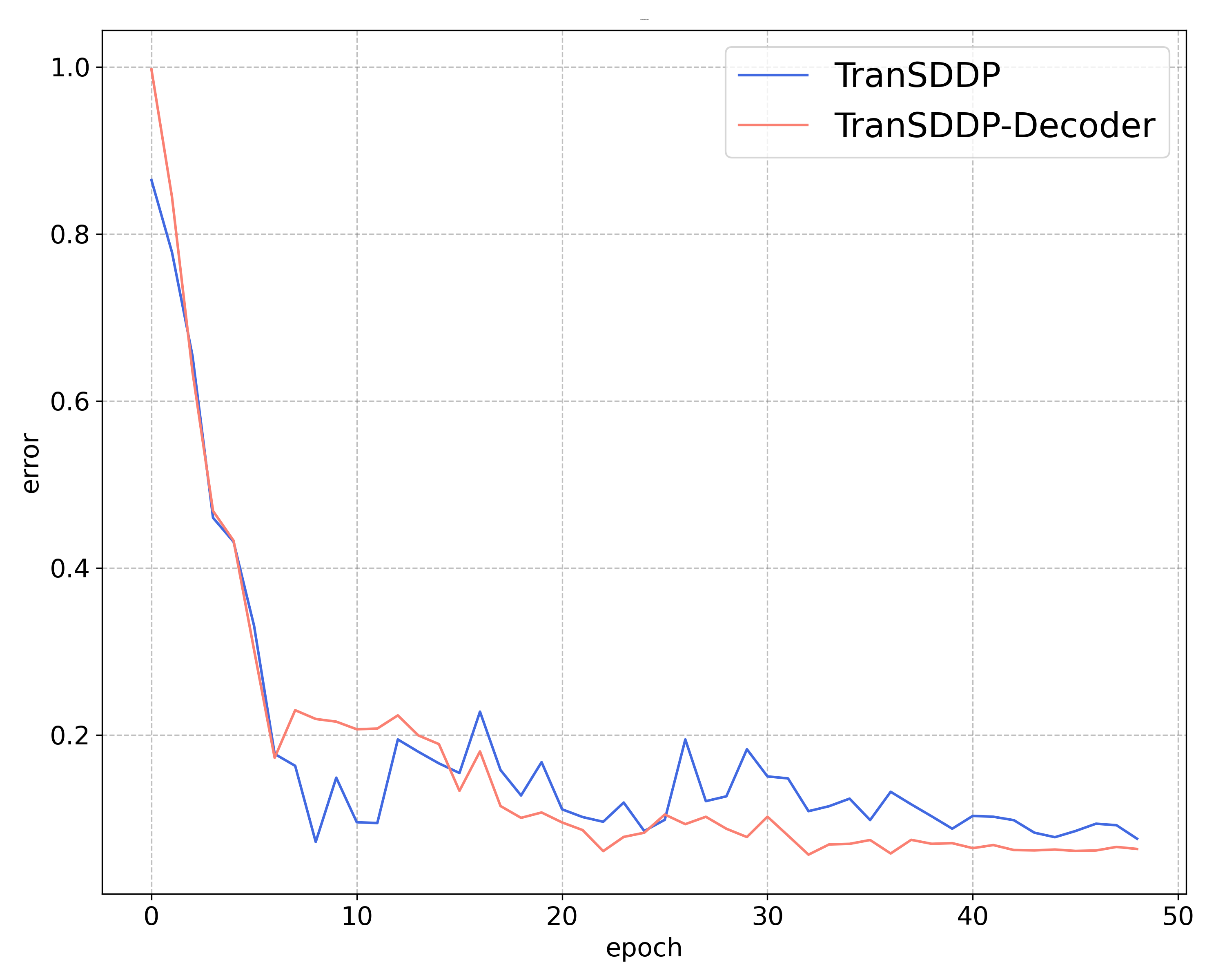}
    }
    \subfigure[Loss]{
    \includegraphics[width=0.4\textwidth]{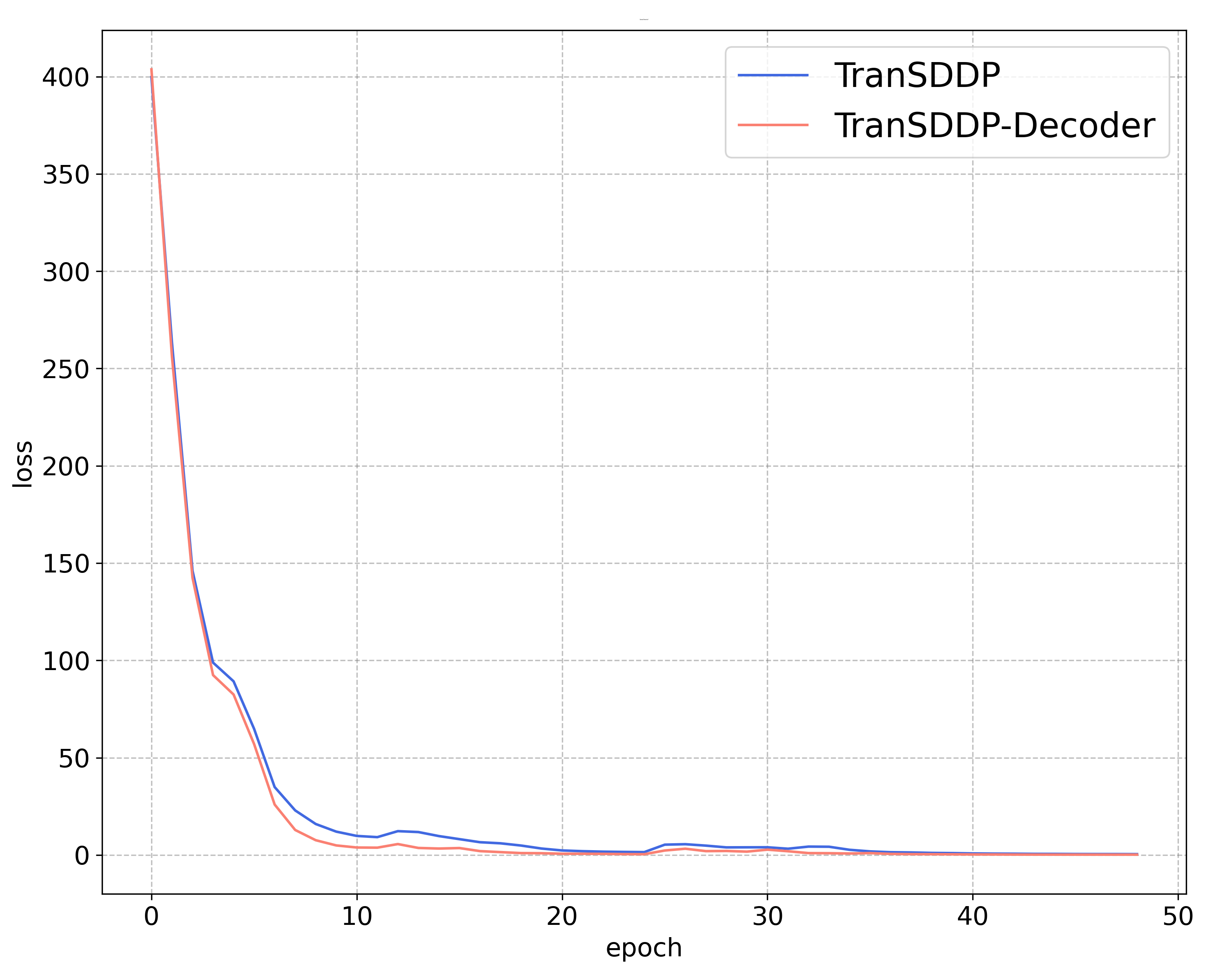}
    }
    \caption{
    Training results for production planning problem
    }
\end{figure}

\begin{figure}[h]
    \centering
    \subfigure[Error ratio]{
    \includegraphics[width=0.4\textwidth]{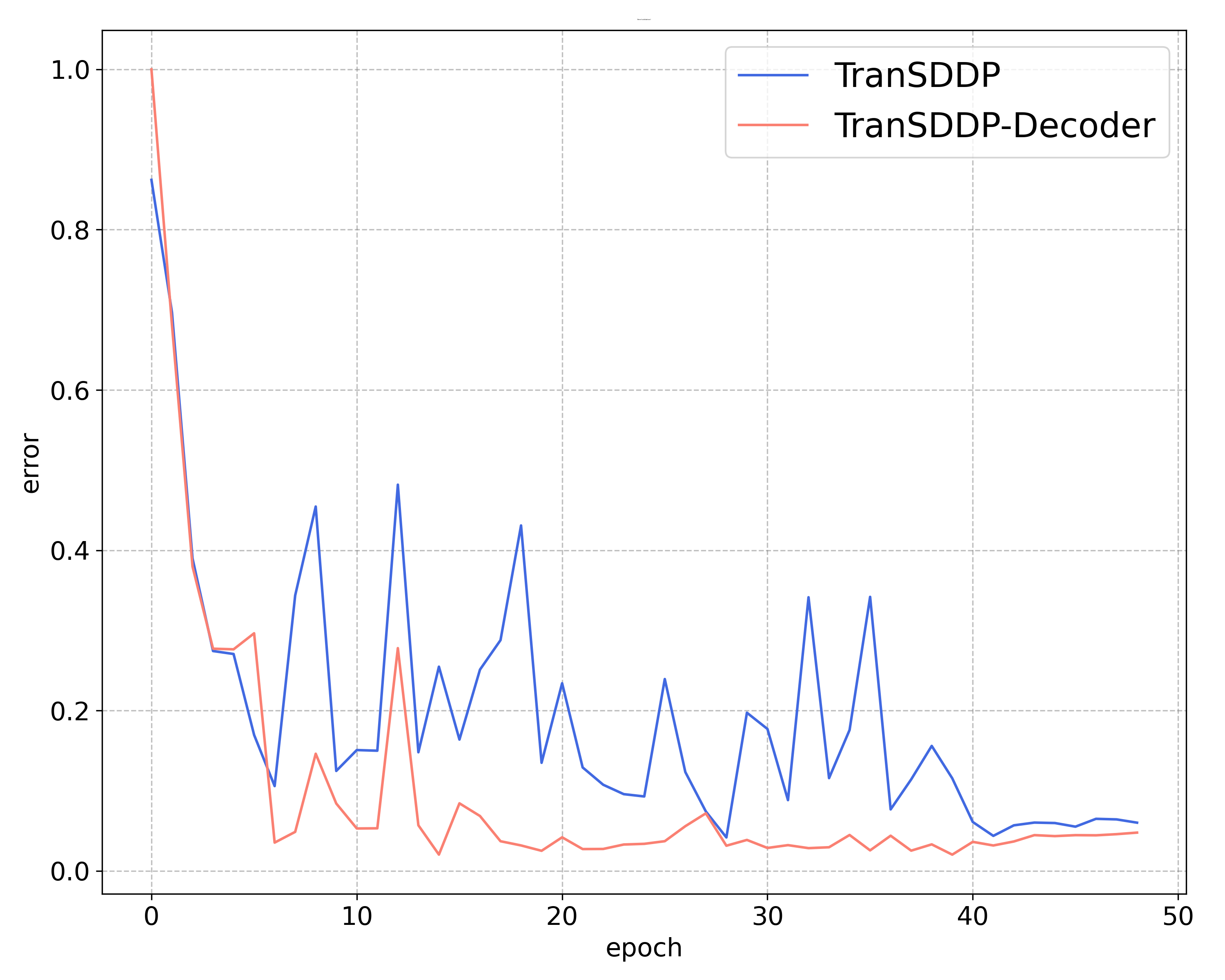}
    }
    \subfigure[Loss]{
    \includegraphics[width=0.4\textwidth]{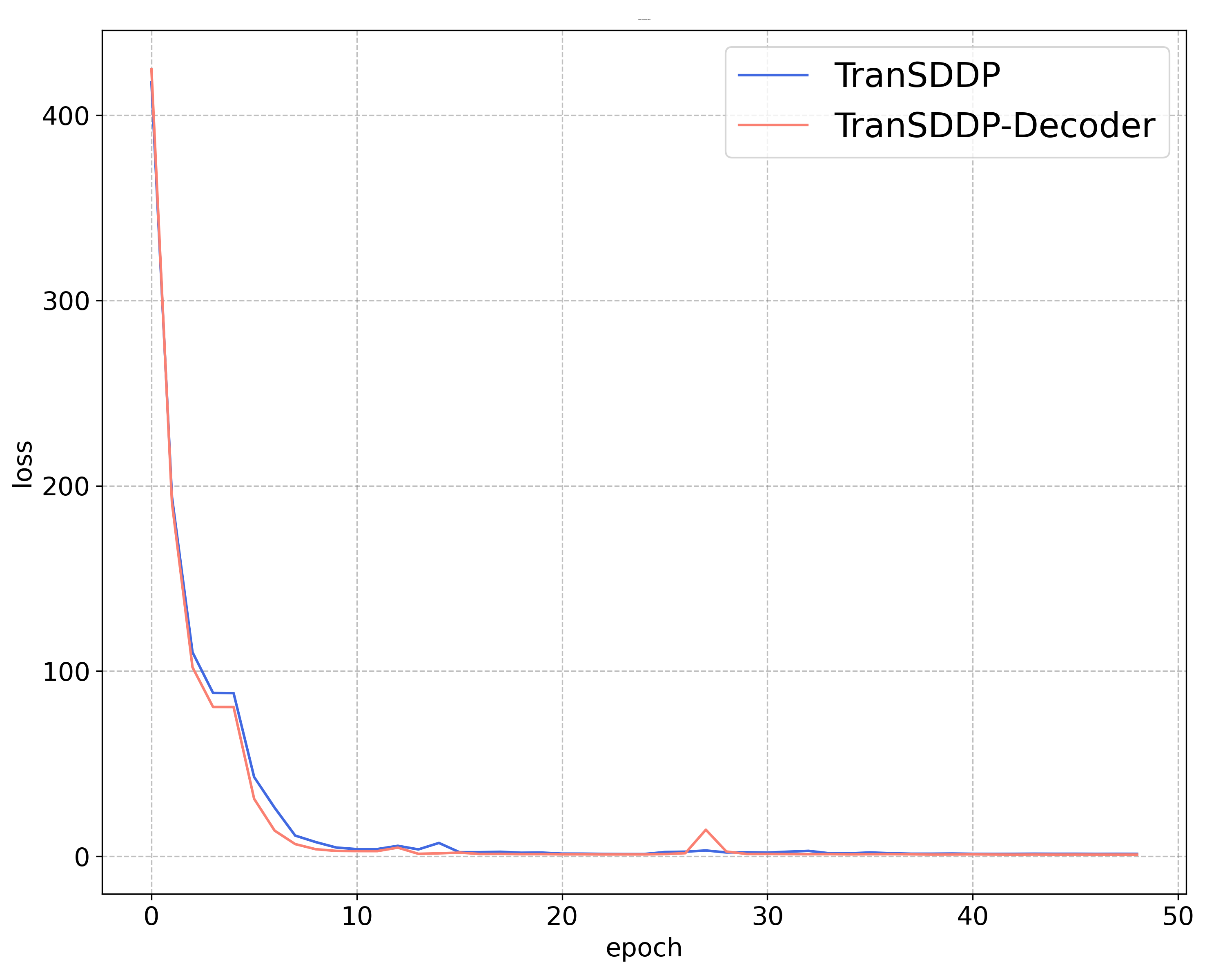}
    }
    \caption{
    Validation results for production planning problem
    }
\end{figure}


\end{document}